\documentclass[journal]{IEEEtran}
\usepackage{amsfonts}
\usepackage{amssymb}
\usepackage{algorithm2e}
\usepackage{algorithmic}
\usepackage{pgfplots}
\usepackage{mathtools}
\usepackage{bbm}
\usepackage{dsfont}
\usepackage{comment}
\usepackage{float}
\usepackage{hyperref}

\ifCLASSINFOpdf
\else
\fi
\begin{document}
\title{CRUR: Coupled-Recurrent Unit for Unification, Conceptualization and Context Capture for Language Representation - A Generalization of Bi Directional LSTM} % Sentence Tensor Product Representation  for

\author{Chiranjib~Sur \\ %,~\IEEEmembership{Member,~IEEE,} 
		Computer \& Information Science \& Engineering Department, University of Florida.\\
		Email: chiranjibsur@gmail.com
        %John~Doe,~\IEEEmembership{Fellow,~OSA,}
        %and~Jane~Doe,~\IEEEmembership{Life~Fellow,~IEEE}% <-this % stops a space
%\thanks{M. Shell was with the Department of Electrical and Computer Engineering, Georgia Institute of Technology, Atlanta, GA, 30332 USA e-mail: (see http://www.michaelshell.org/contact.html).}% <-this % stops a space
%\thanks{This draft is being submitted to the PhD committee members of the author as part of the Qualifying Examination formality of Computer \& Information Science \& Engineering Department, University of Florida.}
%\thanks{Manuscript received August XX, 2016}}%; revised August 26, 2015.}
}


\markboth{Journal of XXXX,~Vol.~XX, No.~X, AXX~20XX}%
{Shell \MakeLowercase{\textit{et al.}}: Bare Demo of IEEEtran.cls for IEEE Journals}

\maketitle

\begin{abstract}
In this work we have analyzed a novel concept of sequential binding based learning capable network based on the coupling of recurrent units with Bayesian prior definition. The coupling structure encodes to generate efficient tensor representations that can be decoded to generate efficient sentences and can describe certain events. These descriptions are derived from structural representations of visual features of images and media. An elaborated study of the different types of coupling recurrent structures are studied and some insights of their performance are provided. Supervised learning performance for natural language processing is judged based on statistical evaluations, however, the truth is perspective, and in this case the qualitative evaluations reveal the real capability of the different architectural strengths and variations. 
Bayesian prior definition of different embedding helps in better characterization of the sentences based on the natural language structure related to parts of speech and other semantic level categorization in a form which is machine interpret-able and inherits the characteristics of the Tensor Representation binding and unbinding based on the mutually orthogonality. Our approach has surpassed some of the existing basic works related to image captioning. 
\end{abstract} % Product
%We have applied the procedures with image caption generation.  
%and video caption generation applications. 

 % product

% \fi
\IEEEpeerreviewmaketitle

\section{Introduction} \label{section:introduction}
\IEEEPARstart{L}{ong} short term (LSTM) memories are widely analyzed due to their high demand in industry to tackle huge volume of unlabeled data, and data analytic technologies greatly rely on them. Mere object detection and manual tagging failed to provide immense details of the activities and the events in the media data and to overcome the confusion created due to perception and language barriers between human interpretation capability and machine interpretation. Image captioning has progressed but slowed down to gain the optimum efficiency and in this work we have analyzed a new architecture that enhances the image captioning problem from visual features. In disguise, we introduced an effective way of coupling and decoupling tensors which can gather effective representations that can differentiate between different ways of writing and sentence constructions. The new architecture, named Coupled-Recurrent Unit Representation (CRUR) unit, is based on the entanglement of the representation of two recurrent units and passing the knowledge into a form of a structured Tensor Product Representations and decoupling it to the required sentences. 
The main idea behind this architecture is the fact that machine interpretation is based on the fact that machine can only understand orthogonal states and its interpretation can be easily processed and stored. Even, the whole segment of network and channel coding, signal detection and estimation theory is dependent on the orthogonal properties. 
After successful utilization of the Tensor Product Representation based on Hadamard matrix for question answering \cite{palangi2017deep} with high accuracy of prediction for answers, there was a need to informalize, regularize and generalize the concept to learnable representations that is scalable and can take forward the ultimate burden of deciphering huge amount of data and be helpful to mankind. \cite{huang2018tensor} provided some analysis on this architecture, but was limited to LSTM, while this work provided a much elaborated analysis and outperformed their work. 
Through experimentation, we were able to establish the fact that memory networks can learn the theoretical framework not only for better sentence generation applications, but also were able to gather the framework of sentences and thus will immensely help in revolutionizing the style and pattern of individuals and would be felt less like machines.  Also the coupling effects of the units enhance the effectiveness of the memory and the generated sentences are quantitatively and qualitatively better than the individual units. The main reason of the enhancement is the functional ability of the network to generate representation to control the context of the situation and also the structure of the languages and thus effective in learning diversified representation from their product for the languages, a mimicry of human artifact, identity and difference. 

In CRUR model, the information content of the overall structure is enhanced and each part can have a different narrative. The fusion of the narratives will help in better long term memory and better contextual learning.  Coupling of the RNN helps in better retaining capability of the model and the sequential dependency creates better update for the variables and better mathematical model for the architecture. In traditional RNN, initialization was confined to limited sectors, while in CRUR, there are possibilities of dual initialization which can create dual narrative, while there are chances of diverse initialization which can itself be an advantage. However, there are structural differences between the LSTM to enforce non-uniformity and indifference in learning content. One can be regarded as a rule generator and the other is the rule enforcer. However, the basics of rules can never be generalized as they are part of narrative and differs for different samples. The rules narrative is structured internally and theoretically and must not be confused with rule based learning. In language generation, it is generally seen that the RNN structural learning is defined by the prediction of the next probable words than the categorical division. However, since languages are highly structured, researchers claim learning of structures of sentences are also important. But, if the prediction of next word is combined with the next probable rule, it can be more efficient. In fact, we can detect the part-of-speech of the word without knowing the word. 
The most important performance factor for CRUR unit is the effective tensor which represents complex relationships and is sensitive to compression and expansion of the embedding dimension. While expansion can be handled with proper dropout, compression can create limitation in variations of expression and can lead to error in part-of-speech and also shortening of description of events. 
While coupling is linear transformation of matrix, one part can be the driver for words occurrence while the other is for grammatical rules or part-of-speech. Initialization of the word predictor must come with context (like visual features) while the one with grammatical rules must be initialized with rule information (tagging distribution). While data driven training will not provide very stable and generalized part-of-speech representation, the perfection and variation in context depends on initialization and proper coupling of the two representations. 

The rest of the document is arranged with 
problem description of language in Section \ref{section:problem}, 
theories and advent of the tensor product representation in Section \ref{section:theory},
discussion of the role of Bayesian Prior in Tensor Product Representation in Section \ref{section:Bayesian},
architectural details of the different CRUR models in Section \ref{section:CRUR},
methodology details of the application along with details of data in Section \ref{section:methodology},
experiment details, results and analysis details in Section \ref{section:results},
%details of language learning in Section \ref{section:Language},
%language structure prediction in Section  \ref{section:LanguagePrediction},
%language structure controlling in Section \ref{section:LanguageControling},
revisit of the existing works in the literature in Section \ref{section:literature},
conclusion and discussion in Section \ref{section:discussion}. 

Our main contribution consists of the followings: 1) novel architecture for sentence representations where context features and language attribute feature cooperate for sentence generation 2) enhanced performance for architectures with just image features, achieved a BLEU\_4 value of 32.7\% in comparison to the previous works 3) logical establishment of the mathematical modeling for tensor products, more than sequential establishments of bi-directional architecture 4) inclusion of language attribute influence and their representation for generation of sentences 5) enhanced predictive language attribute modeling from $\textbf{u}_t$, before decoding the words from $\textbf{f}_t$ with very high accuracy 6) language attributes based controlled analysis for generation of complex and compound sentences.

%{\color{red} Works left : Literature Review}

\begin{figure*}[!h]
\centering
\includegraphics[width=.95\textwidth]{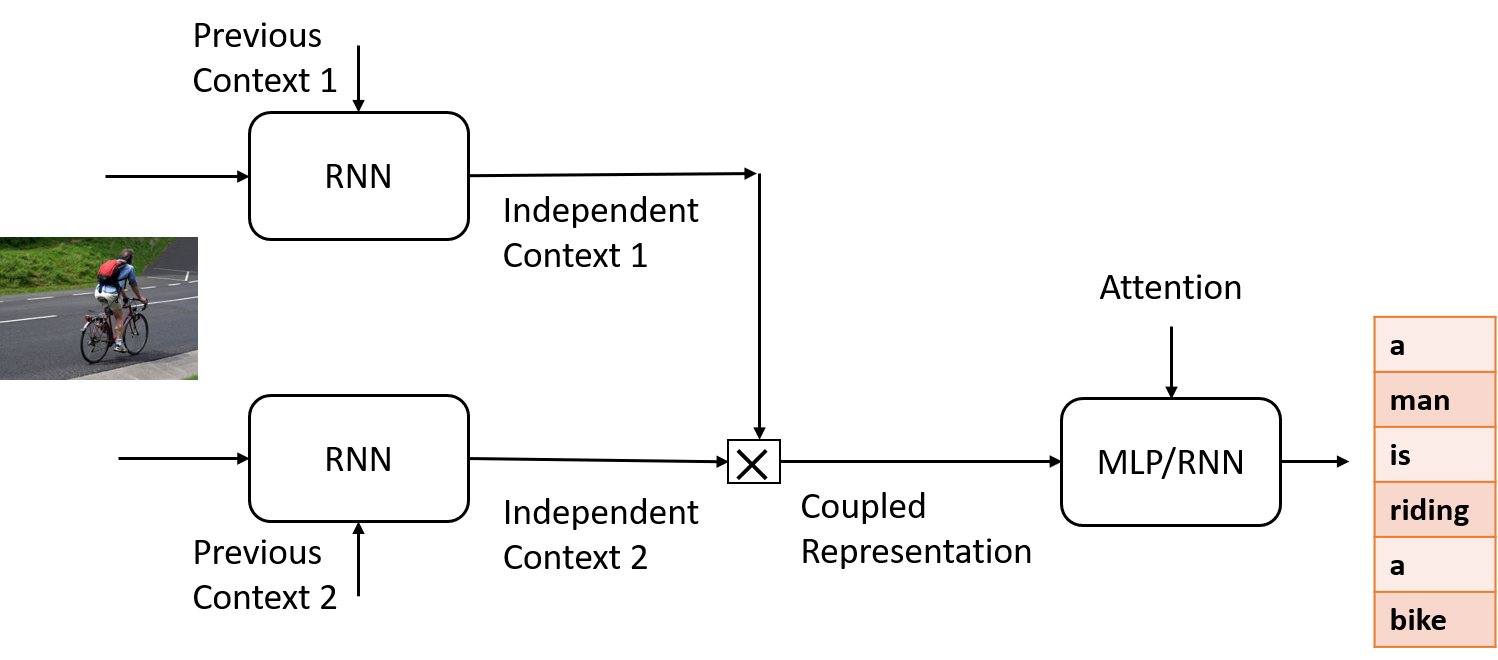}  % ,height=\textheight,keepaspectratio
\caption{Generalized Model of Coupled-Recurrent Unit Representation (CRUR) Unit. Bi-Directional LSTM is a  special cases of CRUR for coupling sequences for classification, while image captioning is generation based application.} \label{fig:Generalized_Model.png}
\end{figure*}

\begin{figure*}[!h]
\centering
\includegraphics[width=\textwidth]{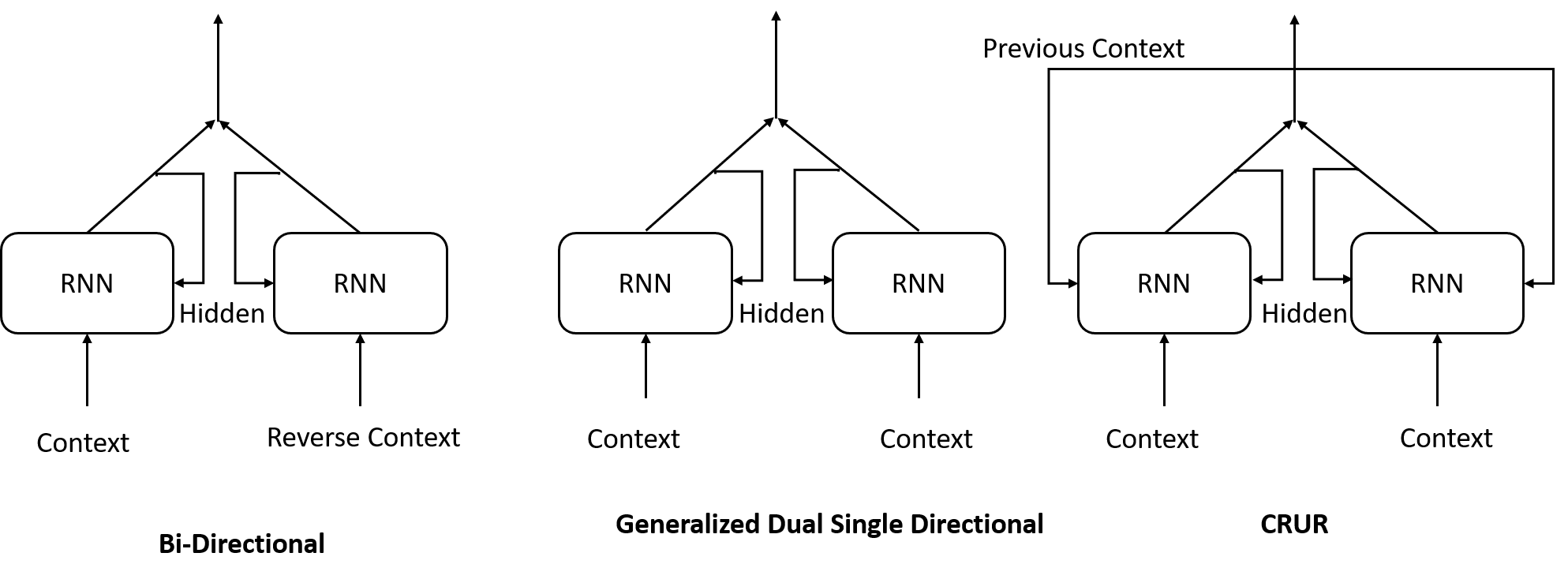}  % ,height=\textheight,keepaspectratio
\caption{Generalized Model of Coupled-Recurrent Unit Representation (CRUR) Unit and its Differences with Bi-Directional LSTM.} \label{fig:Generalized_Model_Differnces.png}
\end{figure*}

\begin{figure*}[!h]
\centering
\includegraphics[width=\textwidth]{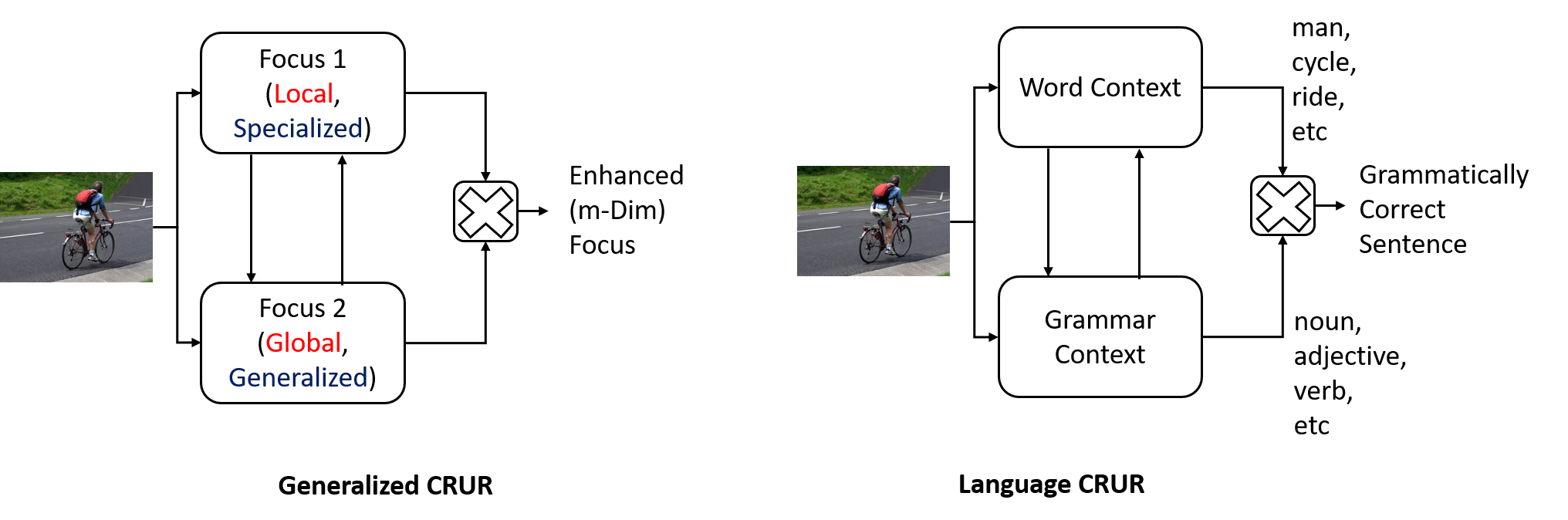}  % ,height=\textheight,keepaspectratio
\caption{Coupled-Recurrent Unit Representation (CRUR) Unit Focuses on Two Different Aspects From Source Like Local-Global, Specialized-Generalized, Word-Grammer etc. The Contexts Vary From Application to Application.} \label{fig:Generalized_Model_Expl.png}
\end{figure*}

\section{Problem of Language Generation}  \label{section:problem}
Machine's ability to write based on events, contexts and facts, creates problem for many applications to directly convey their diverse messages to the users, the problem which was tackled through finite set of indications. Machines must come out of these finiteness with capability of generating texts from contexts and an infinite range of topics through different languages. While, large part of these non-deformed contexts comes from visual representations like images and videos, narrative contexts are associated with uncertainties and ambiguities for machines for inference. Ambiguities in visual contexts are relatively less, but the number of prospective increases. In this work, we have provided a new concept for generation of texts through the utilization of language attributes like parts-of-speech, semantics etc. In absence of concrete and generalized language rules, mathematically defining advanced models and data driven techniques are the best way to learn sentence construction. In this work, we have provided some instances of the approximation of the concepts through establishment of the topological relationships among different words. Generation of texts has many applications and each individual application is defined by different model approximation and is obsessed with their own problems. 

\subsection{Text Generation Problem}
Text Generation is the ability of computer to generate texts from contexts like human beings sensing certain conditions and state space of the system. However, due to immense confusion in representations and structural differences in defining the functional approximation of the model, image captioning application found immense struggle in gaining high momentum for the considered statistical evaluations. However, if unique representation is produced and the models can be made sensitive to the minute variations in the representations, then caption generation can be made very effective. However, the irony is that deep learning deals with suppression of the variations and thus merges the samples to a distribution generated during the training sessions. Mathematically, we can define Text Generation model as $\{w_1,\ldots,w_n\} = f(T,\textbf{W})$ where $T = f(\textbf{v})$ with $w_i$ represent words, $T$ is the context representation, $\textbf{v}$ context features, $\textbf{W}$ are the estimated parameters. Text generation faces lots of problems including the biasness of memory networks  to gather similar kinds of words after certain known words and lack of diversity in generation. This problem is generated from the suppression of variation of the representations, where the network conducts approximations through non-linear transformation of the contexts.  However, the network should focus on suppression of variation of the individual features and that will help in determination of effective representation that can help in caption generations. 
The main cause of this problem is also the lack of definition of proper representation that can help in diversification of predictions and help in generation of words in sentences that were never used in training. Representation, mostly used, are something that is working, and not defined involving the existing embedding of word vectors. 

\subsection{Reply Generation}
Reply Generation is the contextual generation of texts based on the textual query on a conversation. It is also a very important application of modern day world as the interaction between human and machine is not limited to one-way interception but two-way conversation. It requires both the agents interact with each other in a shared common language and understand what the other is saying. 
Mathematically, we can define Reply Generation problem as generating related sentences $\{w_1,\ldots,w_n\} = f(T,\textbf{W})$ where representation $T = f(\{p_1,\ldots,p_m\}) \subset f(\{q_1,\ldots,q_k\})$ is dependent on the query (context) series $\{q_1,\ldots,q_k\}$ being made in the interaction series $\{p_1,\ldots,p_m\}$. 

\subsection{Language \& Style}
Language differs in style and with change in style, the pattern of appearance of parts-of-speech (POS) changes. While we define an architecture, which can even provide a structural component for POS, it is eminent that the memory network is learning to generate sentences, which can be regarded as simple, complex and compound sentences based on the number of independent and dependent clause it contains. An independent clause can be regarded as a sentence, representing a complete thought, while dependent clause even though it has a subject and a verb, cannot be regarded as a sentence. Simple sentences would have one independent clause that is one subject and one predicate, while Complex sentences contain an independent clause and at least one dependent clause. Compound sentences has at least two joined independent clauses. The machine must learn these clauses and learn how to use them in the flow of conversation and sentence generation with logical construction of the action and activities.  

Other aspects of the sentence construction are the language attributes that linguists have recognized through ages and classified the pattern, however failed to provide some concrete set of rules and algorithms for generation. This is where our work focused on to make machine learn the taste of aesthetic writing specific to disciplines. Knowledge of the grammar and part of speech is efficient but lack of concrete rules makes it very difficult for the machine to learn, while prediction memory model can predict easily based on the likelihood. While most of the language models are based on prediction of the next word, the likelihood of decision is dependent on one set of estimated parameters. In this work, we have utilized two models dedicated for context sequentiality and construction topology for word selection and parts-of-speech respectively.
Also, while predicting the sentence, what we expect from the model to learn unique representations of the words and also the structural interpretation of the sentences and emphasis on machine comprehension through prediction models and later demonstrate to control the generation.
Mathematically, we can define $\{w_1,\ldots,w_n\} = f(T,\textbf{W}, S)$ where $S$ is the style factor related to language attributes.

\section{Theory of Tensor Product}  \label{section:theory}
A new tensor called Tensor Product $T$ is generated  through the multiplication of two set of tensor series (filler $\textbf{f}$ and binder $\textbf{r}$) with prescribed interpretation and one set of tensor $\textbf{r}$ can help in perfect regeneration of the other $\textbf{f}$ from the new tensor $T$. Here $\textbf{f}$ is related to context interpretation and $\textbf{r}$ to language attribute interpretation and the combined give rise to the word level accuracy. Tensor Product and their generated Representation uses the concept of linearly independence with transformation and inverse transformation on the assumption that the inner product will help in localization. 
However, as large part of real world scenario and their associated problems are variational, the linearly independence criteria can be relaxed and represent the tensor product with other semi-independent vectors and rely on the assumption that tensors are far apart to interfere and the high dimension of $\textbf{f}$ (or if needed $\textbf{r}$) will provide adequate independence space for each of them. Generalized TPR can be represented as $\textbf{s}(\textbf{w})$ as, 
\begin{equation}
	\textbf{s}(\textbf{w}) = \sum f_i \otimes r^T_i
\end{equation}
where $\textbf{w}$ is the feature vector, and $\{\textbf{w} \rightarrow \textbf{f} : \textbf{w} \in \textbf{W}_{e} \}$ is the transformation, $\textbf{W}_{e}$ is the raw features or the embedding vectors for features which minimizes the context function such as Word2Vec for $k$ contexts as $W2V\_Fn = \min \sum\limits_{i} \sum\limits_{j = 1}^{k} ||\textbf{w}_i - \textbf{w}_j||^2 $, $\textbf{r}$ is the independence imposer for the TPR. This kind of tensor product creates a combined representations of the whole feature space and is  unique, reduced in dimension and with the following decoupling equation,
\begin{equation}
	\textbf{f} = \textbf{s}(\textbf{w}) \otimes \textbf{r} = \textbf{s} \otimes \textbf{r}
\end{equation}
where we can generate back the complete original vector $\textbf{w}$ from $\{\textbf{f} \rightarrow \textbf{w} : \min \limits_{\forall i \in N} \arg  (\{\textbf{f}_i\} - \textbf{f}) \}$ without any error. We have $N$ sample instances and $\min \limits_{\forall i \in N} \arg (\{\textbf{f}_i\} - \textbf{f})$ points to the closest possible sample $i$.  So what we can conclude that this $\textbf{s}(\textbf{w})$ instantiate a much better comprehensive and compressed state of the samples than the whole feature space and can be help many learning algorithms to create models that can understand and differentiate the representations without explicitly supervising it to learn that these are different and need to be differentiated. At the same time, the feature representation can be migrated to its original form in constant time. Previous approaches for transferring 
$\min \limits_{\forall i \in N} \arg (\{\textbf{f}_i\} - \textbf{f})$ point to the closest possible sample $i \in N$ were cosine distances or nearest neighbor with distance norm. However, the same task is possible in constant time as a transformation $\max\limits_{\forall i \in N} \arg \textbf{W}_f \textbf{f} = \textbf{f}_i$ through posing the problem as a probability distribution as we tune our model to gradient error rectification and learning schemes.

\subsection{Tensor Product History}
Tensor Product has been widely used in signal processing and other applications related to mixing of signals and other network coding and channel coding applications where data packets are transmitted using the tensor product concept to retrieve at the other end. Wireless network coding also uses orthogonality coupling for transmission of data packets. 
Spectrum detection theory uses this orthogonality concept widely for many applications for mixing of the signals based on different basis and later uses the property of tensor product for detection and estimation. Here each basis consists of one representation and the combination of these representation create a higher level representation for others. The spectrum theory principle is based on the assumption that the basis $\textbf{b}_{p_i}$ tensors are orthogonal (orthonormal) or $p_i$ phase-shifted and when the signals $\textbf{s}_i$ are multiplied with these vector, they ($\textbf{D} = \sum \textbf{b}_{p_i} \textbf{s}_i$) are also phase-shifted and at the detection part, the multiplication of the same basis revert back the original signal $\textbf{s}'_i = \textbf{D} \textbf{b}_{p_i}  \approx  \textbf{s}_i $ without noise. The equations that explains this phenomenon are provided as follows. 
\begin{equation}
 \textbf{d}_i = \textbf{b}_{p_i} \textbf{s}_i
\end{equation}
\begin{equation}
 \textbf{D} = \sum \textbf{d}_i = \sum \textbf{b}_{p_i} \textbf{s}_i
\end{equation}
\begin{equation}
 \textbf{s}'_i = \textbf{D} \textbf{b}_{p_i} 
\end{equation}
where we have $\textbf{s}_i$ as the original signal, $\textbf{b}_{p_i}$ is the basis transformer, $\textbf{d}_i$ is the transformed signal with phase shift and $\textbf{D}$ is the combined signal to be transmitted. $\textbf{s}'_i$ is the regenerated signal at the reception end. 
$\textbf{b}_{p_i}$ posses the property of orthogonal basis function like Cosine, Sine, etc.

\subsection{Unification of Symbolism and Naturalism}
Unification of Symbolism is marked by generation of global representation for symbols and these representations have the capability to learn the intricacies and rules of the operational procedures on the symbols. For example, in case of natural language processing, unification of symbolism is denoted by the capability to represent alphabets with orthogonal one-hot vectors and by continuous representations like GloVe, and tensor product representation can learn the grammatical rules of sentences through the use another vector that can be denoted as the dictator of the next probable parts-of-speech. These dictators will never be perfect for memory network based prediction due to the fact that the whole notion in memories is approximate representation and this is done to scale up the learning capabilities. 

Naturalism is an important criteria for sentence generation and is a way to prevent language construction biasness. Construction biasness is defined as the appearance of similar kind sentence rules for sentence and the machine capability to learn only limited way of expressing themselves. This problem of construction biasness in machine generated sentences is known as the problem of Naturalism and need to be dealt with as we move towards more sophisticated systems and capability to generate meaningful sentences from contexts. 

\subsection{Hadamard Matrix}
Generation of many orthogonal structures is difficult and hence Hadamard Matrix is used for initialization. 
Hadamard Matrix consists of series of orthogonal rows and columns and its generation ensures such functionality. While dealing with TPR generation and other prediction and detection frameworks, generation and maintenance of the mathematical constraints becomes important for the best performance. 
In general, Hadamard matrix is a $(2^n \times 2^n)$ square matrix, consisting of $\{-1, 1\}$ and each of the rows are orthogonal to all the others. The consequence is that, it can be used to generate mutually independent vectors for the TPRs. Hadamard Code TPR was build on top of Hadamard Coded matrix using the following equations. 
\begin{equation} \label{eq:hadamard}
H_{2^n} = \frac{1}{c_{k-1}}
\left[{\begin{array}{cc} 
		H_{2^{n-1}} & H_{2^{n-1}} \\ 
		H_{2^{n-1}} & -H_{2^{n-1}}
		\end{array}}\right]
= \frac{1}{c_{k-1}} H_{2} \otimes H_{2^{n-1}}
\end{equation}
$H_{2^n} \in \mathbb{R}^{nnnn \times nnn}_{0/1}$ mostly consists of zeros and ones. The rows and columns of $H_{2} \otimes H_{2^{n-1}}$ are symmetric and form bases of Hadamard matrix where we have $\otimes$ as the Kronecker product, $\frac{1}{c_{k-1}}$ is the normalization factor where $c_{k-1} = ({\sum |x_{i}|^2})^{\frac{1}{2}}$ with Frobenius norm or $L^2-$norm of any row as the normalizing coefficient. If we consider $(k-1) = 2$, then the most fundamental Hadamard matrix with $c_{2} = c_{(k-1) = 2} = ({\sum |x_{i}|^2})^{\frac{1}{2}}$ is denoted as the following, 
\begin{equation} 
	H_{2} = \{ \frac{1}{c_{2}} \}
\left[{\begin{array}{cc} 
		1 & 1 \\ 
		1 & -1
		\end{array}}\right]
		  = 
\left[{\begin{array}{cc} 
		0.707 & 0.707 \\ 
		0.707 & -0.707
		\end{array}}\right]
\end{equation} 
This matix forms the starting matrix for all other high dimensional Hadamard matrix generation.  
In Hadamard Coding, the filler consists of multiplication of the Hadamard matrix row (Equation \ref{eq:hadamard}) $r^T_i = (H_{2^n})_{i,}$ and the individual feature representations $f_i = (\textbf{W})_{i}$ from feature space $\textbf{W}$ like in case of natural languages, $f_i = (\textbf{W}_e)_{w_k}$ is the word embedding vector for corresponding word $w_k$ from $\textbf{W}_e$.

The Hadamard Code TPR individual is generated as an inner product of the rows $\{(H_{2^n})_{r_i}:i\in \{1,\ldots,p\}\}$ of $p-$level Hadamard matrix (with $\{1,p\}$ dimension) and $ \{F_j:j\in \{1,\ldots,\lceil \frac{d}{q} \rceil\}\} $, the corresponding segment vector (with $\{q,1\}$ dimension) of the $d-$dimensional features of the samples as denoted by $\{F_j\} \{(H_{2^n})_{r_i}\}^T$ to generate a $\{q,p\}$ matrix. Essentially we have $p=2^k$, $p \geq \lceil \frac{d}{q} \rceil$ and symbolically $F = f(\textbf{w})$. 
So the overall Hadamard Code TPR is denoted as, 
\begin{equation}
	\textbf{s}_{H}(\textbf{w}) = \sum\limits_{i} \{F_i\} \otimes \{(H_{2^n})_{r_i}\}^T
\end{equation}
where we can generate back the features $\textbf{w}$ as $\textbf{F}\rightarrow \textbf{w}$ and, 
\begin{equation}
	\textbf{F} = \textbf{s}_{H} \otimes (H_{2^n})
\end{equation}
This procedure helps in the easiest and efficient way of generating and dealing with tensor product representation through linear transformation of the weighted representations of the original features to the mutual orthogonal spaces. Also, TPRs ($\textbf{s}_{H}$), generated from this procedure, have very distinct, non-overlapping and unique feature space for the samples. This created discrete learning phenomenon, which, sometimes, goes against the variation tolerance and regularization compatible network based training models. In such models, connectedness and relatedness, how insignificant they may be, are inevitable part of the learning. This is why, directly dealing with Hadamard Code TPR may not help and there are some extra procedural requirement for the system to work. Next, we will describe in details the procedural flow for image captioning applications, mainly dealing with natural languages. 

%\subsubsection{xxxx}
Let we have sentence with word $w_1,\ldots,w_n$ and word embedding $\textbf{W}_e \in \mathbb{R}^{v \times e}$, we can transfer one hot vector for each word $w_i$ as $(\textbf{W}_e)_i \in \mathbb{R}^{1 \times e}$, we have, 
\begin{equation}
 \textbf{s}_{H} = \sum (\textbf{W}_e)_i * r_j 
\end{equation}
for $w_j = i$ and $\textbf{s}_{H}$ is the TPR.
Conversely, to retrieve the information from the TPR, for each $j \in N$, we have,  
\begin{equation}
  (w_p)_{j} = \textbf{s}_{H} * r_j^T 
\end{equation}  
and if we consider the nearest neighbor for $(w_p)_{j} $ in $\textbf{W}_e$, we find that 
\begin{multline}
 (w_p)_{j} = \arg \min\limits_{k} \{ (\textbf{W}_e)_k \mid \min ||(\textbf{W}_e)_k - (w_p)_{j}|| \} =  \\
 (\textbf{W}_e)_{k=i}  = (\textbf{W}_e)_{i} = w_{j}
\end{multline}
We have tested that the retrieval rate is 100\% correct for word embedding like Word2Vec, GloVe for any dimension. The accuracy of the retrieval is not because of the dimension or the embedding, but due to the mutual orthogonal matrix which creates space for real $f_i$ to be segregated when $r_i$ is multiplied with $f_i r_i^T$ as  $f_i r_i^T r_i$.

\subsection{Experimental Framework for Hadamard Matrix}
Normalized Hadamard Matrix as $\textbf{r}^T$ is used to couple the word embedding ($\textbf{f}$) of words to create TPR and then this TPR is used to generate the words using $\textbf{r}$ through 2-norm nearest neighbor estimation of the generated embedding with the embedding-to-word dictionary. The estimated sequence of words were generated with 100\% accuracy for the training dataset. We tried to map the image embedding with TPR through deep MLP and this MLP can  estimate the training data with high precision, but for testing data, the scheme failed because of the high sensitivity of the model for variations in estimation of  the TPR. However, if we use a nearest neighbor model for  TPR, where the generated TPR for the test data is taken replaced by an already established TPR, we can perform much better accuracy for the test data. If only the training data TPR are used, the accuracy can reach at around 70\% for BLEU\_4 accuracy while if the testing data TPR is used then the accuracy can reach above 90\% BLEU\_4 accuracy.  However, nearest neighbor estimation is time and resource consuming and hence not a prefered solution for modern day applications. Also, nearest neighbor based systems will destroy the notion of generalization of representations and will prevent production of new representations that will create new set of word sequence. Overall, nearest neighbor based solutions do not provide scalability solutions for languages, where the scope of diversity of representations is practically infinite.

\subsection{Approximation, Structuring \& Mutual Orthogonal Problem}
While pure Hadamard matrix row based encoding is sensitive to variations of image features and not scalable, we used LSTM based encoder and decoder. The memeory based models can be initialized and the end-to-end model can generate the perfect intermediate TPR which can be considered as a perfect approximation of representation, that is learned with time and back propagated feedback. The structuring of such representation is generated with the help of an approximately orthogonal vector $\textbf{r}$ and the language attribute $\textbf{f}$ which is associated with the language embedding. The whole system is based on approximation and the coupling and decoupling is deterministic approximation instead of deterministic overall. This assumption and phenomenon worked well for our experiments and had been found to produce better sentences.

\section{Bayesian Prior} \label{section:Bayesian}
Bayesian Prior estimation helps in better modeling and prediction, where the data is represented by a distribution or series of distribution, already estimated or known and the work of the  model is to fit the distribution. Traditional statistical model already assume some kind of distributions for the independent variables and thus facilitates the effectiveness of prediction. But the problem becomes non-trivial when the optimization is multi-modal and best possible solution is not adding much to prediction due to inappropriate estimation of some distributions, data is non-linear though considered to be linear and so on. This problem of estimation of the distribution is caused due to the lack of transformation and processing of the data features, which required to be tamed for better estimation. This is where the deep network helps, where series of layers creates non-linear transformation and approximation of the feature sets to define a much finer and stable generator distribution. Our architecture estimates Bayesian Prior for both contexts $(\mathcal{P}(\textbf{f}))$ and language attributes $(\mathcal{P}(\textbf{r}))$. The TPR $(\mathcal{P}(\textbf{s}(\textbf{w})) = \mathcal{P}(\textbf{f})\mathcal{P}(\textbf{r}))$ is a joint Bayesian Prior generated from their product. Relative variations of $(\mathcal{P}(\textbf{f}))$ and $(\mathcal{P}(\textbf{f}))$ helps in better representation learning. 

We define Bayesian Representation for TPR as an orthogonal set of variable representations that helps in better prediction. These representations can also be regarded as likelihood of contexts and events with sentence composition characteristics. What orthogonality adds to these likelihood is the main point of discussion. While tensor multiplication transforms the feature space to other subspace without significantly judging whether that is beneficial or not, involvement of a orthogonal space creates a discretization and prevents mixing of the different features and thus create enough opportunity to be segregated in the decoder, a phenomenon widely used in signal systems and spectrum detection theory. However, unfortunately there is no way to estimate the inter-working intricacies of the memory network except evaluation at the likelihood level. Moreover, we evaluate on the collective composition than individualism, unlike spectrum detection applications. However, we still work in that direction to establish the principle through estimation and analysis of the vector $\textbf{r} = \textbf{u}_t$ generated through Equation \ref{eq:ut}. It has been proved that $\textbf{u}_t$ is helpful in full prediction of the language attributes and also helps in better estimation and diversification of $f_t$. In analysis we established that the relative orthogonality of $\textbf{u}_t$ vectors for instances of the sentence is beneficial and outperforms previous performances. 

%\textbf{Idea: Bayesian initialization, dropout to visual features and tagging, - this part is not yet included - Bayesian as orthogonal representation}

\subsection{Representation Learning as Bayesian Prior}
Representation Learning as Bayesian Prior is an abstract concept to maintain the feature space and orthogonality will help in maintaining integrity and individuality of the feature space in the hope that it will help in segregate of the features during decoding and decoupling. 
So apparently, out of two set of vectors, $\textbf{S}_t$ will denote the likelihood for a contextual representation to be generated and $\textbf{u}_t$ will the likelihood that it belongs to a certain parts-of-speech or any other language attribute. Combined,  they can help in establishing the likelihood of the word. 
This is the reason why, as a design constrain, $\textbf{S}_t$ is established as a large vector comparison to the smaller vector $\textbf{u}_t$ and to prevent $\textbf{u}_t$ to learn about the contextual representation. It must also be mentioned that representation, as Bayesian prior, aims at providing the best likelihood of the words to be generated and compose the sentence. This is perhaps way different from the end-to-end models where the intermediate representation is far more than a prior of individual likelihood as it needs to generate a series of interconnected sequence likelihood.

\subsection{Tensor Binder as Bayesian Prior} 
Deep learning has always been considered as a system which can approximate the prior estimation from the data based on the  likelihood of the classes. This is the reason, in many cases, the amount of the data is important for establishment of the variation, whereas the prior estimation can be handled through repetition of similar data. 
Tensor binder or the orthogonal vector representation $\textbf{u}_t$ is such an approximation which works on the principle of gathering certain characteristics of sequential connection or topological dependency and can be regarded as Bayesian Prior. The main task of the tensor binder is to gather information related to the possibilities of a context and channelize the system with the best possibility. Like say a context of a `person' can channelize it to `man' or `woman' or `boy', but tensor binder will establish that the word should represent as a noun. 
The relationship for generation of a grammatically correct sentence is prediction of the next context representation and the parts-of-speech representation. Mathematical, we can define, 
\begin{equation}
 \mathcal{P}(R) = \mathcal{P}(X)\mathcal{P}(Y)
\end{equation}
where $\mathcal{P}(X)$ is context representation from image, $\mathcal{P}(Y)$ is the moderator or parts-of-speech representation, $\mathcal{P}(R)$ is the probability of the next event word representation as a noun and a word (say 'man'). This joint influence $\textbf{f}_t$ creates space for new interpretation and can be used to guide grammatically correct sentences and thus free from the short term memory of one LSTM. Coordination and collaboration of the two LSTMs improve performance. Here, $\mathcal{P}(X)$ helps in deciding the next context for the word, provided $\mathcal{P}(Y)$ helps in deciding the style of writing  judged through language attributes like parts-of-speech etc. 

\subsection{CRUR vs Bi-Directional LSTM}
Bi Directional LSTM (bi-LSTM) shares its architecture with CRUR, but limited to a specialized case which performed well for specific applications and does not hold good prospects for data, where the topological relationships hold immense information for inference. In other words, Bi Directional LSTM can be regarded as a special case of the CRUR architecture. Before we discuss CRUR, there is a need to discuss the bi-LSTM architecture and understand why CRUR cropped up as a generalized architecture and what kind of applications are more suitable for bi-LSTM. 
bi-LSTM was established for better prediction and inference ignoring certain aspects related to sequential relationship and topological significance. It was never defined for establishing unique data representation but to converge large part of the similarly classified entities to similar representation that will define some series of additive distributions. bi-LSTM equations can be denoted as the following,
\begin{equation}
 \textbf{h}_{1,t} = LSTM_1(\textbf{h}_{1,t-1},\textbf{x}_{1,t} \mid \textbf{h}_{1,0}, \textbf{x}_{1,t}=\{w_1,w_2,\ldots,w_n\})
\end{equation}
\begin{equation}
 \textbf{h}_{2,t} = LSTM_2(\textbf{h}_{2,t-1},\textbf{x}_{2,t} \mid \textbf{h}_{1,0}, \textbf{x}_{1,t}=\{w_n,\ldots,w_2,w_1\})
\end{equation}
\begin{equation}
 \textbf{y}_{p,t} = f(\textbf{W}_{f}\overrightarrow{\textbf{h}}, \textbf{W}_{b}\overleftarrow{\textbf{h}}) = \textbf{W}_{f}\textbf{h}_{1,t} + \textbf{W}_{b}\textbf{h}_{2,t}
\end{equation}
where it is assumed that the sequences $\textbf{x}_{1,t}=\{w_1,w_2,\ldots,w_n\}$ and its reverse $\textbf{x}_{1,t}=\{w_n,\ldots,w_2,w_1\}$ will infer similar kind of prediction, whereas the real fact is that the two sequences depict completely different representations and can be trained to infer similar expectations. In fact, bi-LSTM tries to converge all combinations of the sequence $\textbf{x}_{1,t}=\{w_1,w_2,\ldots,w_n\}$ to similar kind of distributions, which will in parallel conceive both the ways equivalently. 

But real world problems are much more complex and this kind of assumption can end up to two different inferences for the LSTMs, which can conflict with each other and end up with wrong inference. However, our defined CRUR model is aimed at defining unique representation for better reciprocity, regeneration of composition and global representation and in that respect, we generalize the representation instead of the generalization of the distribution. 
Mostly, deep learning is known to be efficient because of its capability to suppress of the variations for the representation to converge them to pertinent distribution, but we define our network to suppress the numerical at the feature level so that more stable representations are generated and can accommodate the infinite space of languages. 
Dual direction destroys the notion of uniqueness for TPR and will not be good option for natural language where 'I am' and 'am I' is different and must not be used as a converge for the notion of prediction and generation. 
Mathematically, for generative models we can define CRUR as, 
\begin{equation}
 w_n = f(\textbf{f}_t,\textbf{W},\{w_1,\ldots,w_{n-1}\})
\end{equation}
while we can define bi-LSTM generation as,
\begin{equation}
 w_n = f(\textbf{f}_t,\textbf{W},\{w_1,\ldots,w_{n-1}\},\{w_{n-1},\ldots,w_1\})
\end{equation}
Other fundamental differences are defined in terms of capability of generation, where bi-LSTM is focused on inference while transforming the bi-LSTM to a generative one, we ended up with CRUR. Inialization and inter-cooperation or collaboration had been added advantage to CRUR closed model to avoid independent interpretation. In the next few sections, we will discuss more on the CRUR and the way to predict language attributes and control the generation.
CRUR equations can be denoted as the followings, 
\begin{multline}
 \textbf{h}_{1,t} = LSTM_1(\textbf{h}_{1,t-1},\textbf{h}_{2,t-1},\textbf{x}_{1,t} \mid \textbf{h}_{1,0}, \\
 \textbf{x}_{1,t}=\{w_1,w_2,\ldots,w_n\})
\end{multline}
\begin{multline}
 \textbf{h}_{2,t} = LSTM_2(\textbf{h}_{2,t-1},\textbf{h}_{1,t-1},\textbf{x}_{2,t} \mid \textbf{h}_{1,0}, \\
 \textbf{x}_{1,t}=\{w_1,w_2,\ldots,w_n\})
\end{multline}
\begin{equation}
 \textbf{y}_{p,t} = \textbf{W}_{f} \sigma(\textbf{W}_{f} \textbf{h}_{1,t}) \textbf{h}_{2,t}
\end{equation}
where $\textbf{y}_{p,t}$ is interpreted at instances $t = \{1,2,\ldots,n\}$ as $\{w_1,w_2,\ldots,w_n\}$.

\section{Coupled-Recurrent Unit Representatione}  \label{section:CRUR}
\begin{figure*}[!h]
\centering
\includegraphics[width=.75\textwidth]{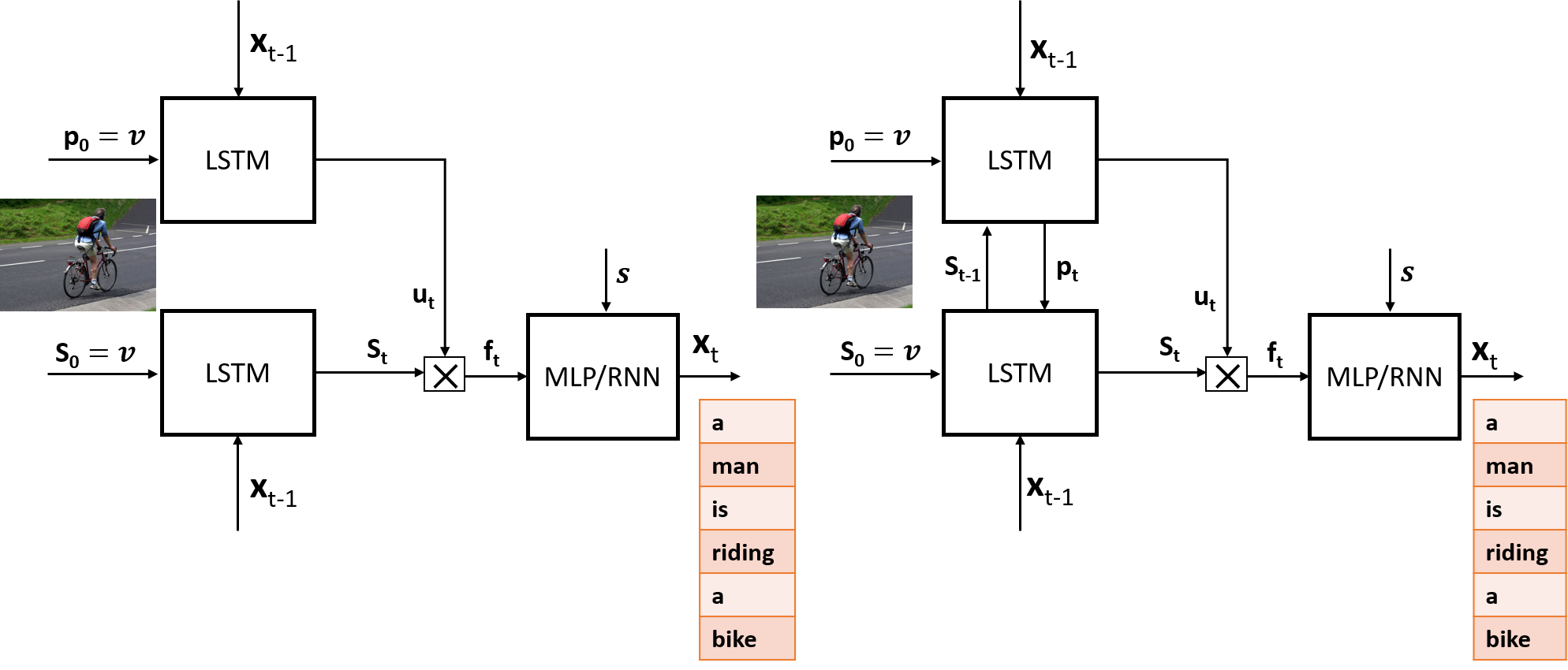}  % ,height=\textheight,keepaspectratio
\caption{Basic Model for TPR Generation (Left formed Open Ended Model,  Right formed Closed Ended Model)} \label{fig:basicTPRa}
\end{figure*}

\begin{figure*}[!h]
\centering
\includegraphics[width=.75\textwidth]{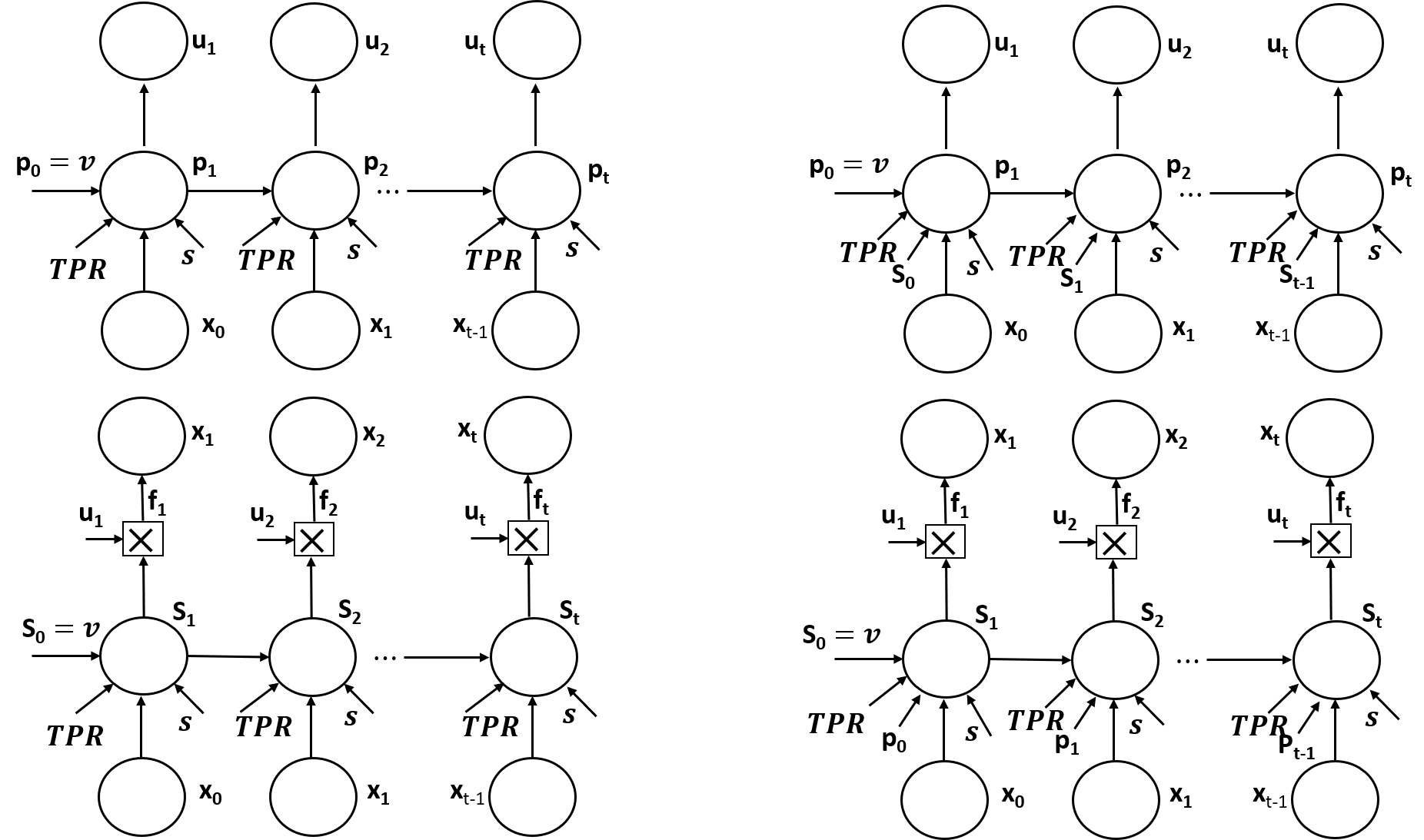}  % ,height=\textheight,keepaspectratio
\caption{Basic Model for TPR Generation (Upper and Lower Left formed Open Ended Model, Upper and Lower Right formed Closed Ended Model)} \label{fig:basicTPRb}
\end{figure*}

The Coupled-Recurrent Unit Representation (CRUR) unit \cite{Sur2018Representation}, \cite{palangi2017deep}, \cite{huang2018tensor} is an entanglement or tensor product of different interpret-able tensors along with crafted initialization of the parameters. The overview of the generalized CRUR architecture has been pictured both in Figure \ref{fig:basicTPRa} and Figure \ref{fig:basicTPRb}. The success of CRUR depends on the hidden state sharing scheme and transfer of knowledge of one LSTM with the other and thus can coordinate and cooperate. That is the reason why, CRUR provided a much better effect than traditional individual LSTM. The main reason of better learning capability of CRUR is the wide range of dependencies and sharing of variables and intermediate states to complement each other and also due to generation of regularized and specialized tensor representations, which drive the architecture to generate better visual captions. Tensor product of different tensors diversifies the opportunity of different combinations of likelihood and prevents the model from learning biases. We will mostly deal with LSTM, but different recurrent units can be used to generate different architectures and we have done elaborated study of some of  them to understand their learning capabilities. 

The main differences of these architectures are based on the amount of interdependence of the state spaces, number of activation gates, which also define the diversity of substances and also the amount of knowledge it can learn. For example, GRU provided better learning due to the large range of simultaneous triggering of the non-linear functions and facilitating the incorporation of knowledge, while LSTM can produce much better sequences. In our new architecture, there are  functionality related to learning generation of sentences and also the ability to interpret different definition of tensor representation. This representation is important as it helps in bringing together different aspects of the languages from visual features to combine and generate. 
Such representations are derived from individual recurrent units and are expected to evolve. 

\subsection{Basic Conceptual Model - Open Ended}
Open Ended models are the most basic model and the bi-LSTM is a special case where the LSTMs are symmetrical in dimension, each is trained with either forward or backward sequence and the ensemble of the likelihood $\textbf{h}_{1,t}$ and $\textbf{h}_{2,t}$ can be generalized with addition $(\textbf{h}_{1,t}+\textbf{h}_{2,t})$, multiplication $(\textbf{h}_{1,t} \odot \textbf{h}_{2,t})$, concatenation $[\textbf{h}_{1,t},\textbf{h}_{2,t}]$ or even weighted combinations $(\textbf{W}_1\textbf{h}_{1,t}+\textbf{W}_2\textbf{h}_{2,t})$ of the two. However, open ended models are widely used for fusion of information and without any sharing of information open ended models tend to have different opinion of the same contextual relationship. This resulted in lower rate of learning, no mutual sharing of knowledge, lower number of variables and also low approximations. Mathematically, we can define Open Ended models as the followings, 
\begin{equation}
 \textbf{h}_{1,t} = LSTM_1(\textbf{h}_{1,t-1},\textbf{x}_{1,t} \mid \textbf{h}_{1,0})
\end{equation}
\begin{equation}
 \textbf{h}_{2,t} = LSTM_2(\textbf{h}_{2,t-1},\textbf{x}_{2,t} \mid \textbf{h}_{2,0})
\end{equation}
\begin{equation}
 \textbf{y}_{p,t} = \textbf{W}_{1}\textbf{h}_{1,t} + \textbf{W}_{2}\textbf{h}_{2,t}
\end{equation}
where $\textbf{h}_*$ is the generated hidden states of the memory network and $\textbf{x}_*$ are the inputs. The open ended models are described with detailed equations in the subsequent subsections. Nevertheless, some applications will find open ended model better due to the fact that traditional feature learning systems tend to perform well when ensemble of extracted feature (like boosting) are used for inference.

%Problems: low rate of learning and no mutual sharing of the knowledge and learning, low variables and hence approximation is also low

\subsection{Tensor Representation Structure \& Size Analysis}
In our model, we have emphasized on asymmetrical LSTM structures for CRUR, which will help in reduction of variable estimations and at the same time will bound the interpretation of the representations and reduce sparsity. However, the dimension of the two LSTMs must be proportion so that the effect of one can be reflected in the other and can bring changes in the likelihood estimation of the sequence. Initially, we considered $\textbf{u}_t \in \mathbb{R}^{10}$ and $\textbf{S}_t \in \mathbb{R}^{10\times 10}$ while tensor product produced was $\textbf{f}_t \in \mathbb{R}^{10}$. Noticeably, $\textbf{f}_t \in \mathbb{R}^{10}$ was not enough for representation and converged the representation to non-variational tensors, although the theoretical framework supported that. We increased our model dimension to $\textbf{u}_t \in \mathbb{R}^{10}$ and $\textbf{S}_t \in \mathbb{R}^{10\times 10}$ while now tensor product provided was $\textbf{f}_t \in \mathbb{R}^{10\times 10}$. 
The first model, with $10 \times (10,10) = 10$, failed to work properly due to two reasons: no end-to-end framework (decoupling is possible there) and the disproportion fact that 10 was too low for the other to mingle around and act as a proper initial representation that can capitulate the image tensor as sentence. 
Later, we changed the dimension to $10 \times (10,10) = (10,10)$ or even $10 \times 10 = 100$ may work for some applications. This architecture performed better and we can structure the captions much more effectively than the previous model. 

%Initial it was $10 \times 10 = 10$

\subsection{Open-End \& Closed-End Schemes for CRUR}
We have defined two architectures for the coupling decoder: one with a Multi-Layer Perceptron called shallow scheme and another with a recurrent unit called deep scheme. Recurrent units are widely accepted because of the end-to-end learning capability. While we compare the different schemes, it is important that we understand that the criteria of learning is both contextual and combination. While many caption strategies go for contexts, their low BLEU value indicate that they fail to generate the combination of visual content. We have provided some these incites when we do the qualitative analysis of the generated captions in Section \ref{subsec:Qualitative}. 
The followings are the probable decoders and not generator and hence $\textbf{f}_t$ is used as attention and not as initialization and for each decouple, it is replaced by a new one. For MLP, the decouple equations are as follows,
\begin{equation}
 \textbf{y}_{t} =   (\arg\max\sigma( \textbf{W}_{x}\textbf{f}_{t} )) 
\end{equation}
While LSTM, these are the decoupling equations for decoding, 
\begin{equation}
 \textbf{y}_{t} =   (\arg\max LSTM( \textbf{x}_{t}, \textbf{f}_{t}, \textbf{h}_{t-1}  = \textbf{h}_{0} )) 
\end{equation}
where $\textbf{h}_{0}$ is initialized with constant. Even this equations may work for some applications.
\begin{equation}
 \textbf{y}_{t} =   (\arg\max LSTM( \textbf{x}_{t}, \textbf{f}_{t}, \textbf{h}_{t-1}  = f(\textbf{f}_{t}) )) 
\end{equation}
and is free from initialization. It must be mentioned that a LSTM decoder will be more sensitive and can differentiate between more among the visual features than the MLP layer and hence a default choice for many applications. Sensitive means that LSTM can facilitate more diverse caption generation and will help in segregation of more number of features to appear in the generated sentences.
\begin{figure}[H]
\centering
\includegraphics[width=.5\textwidth]{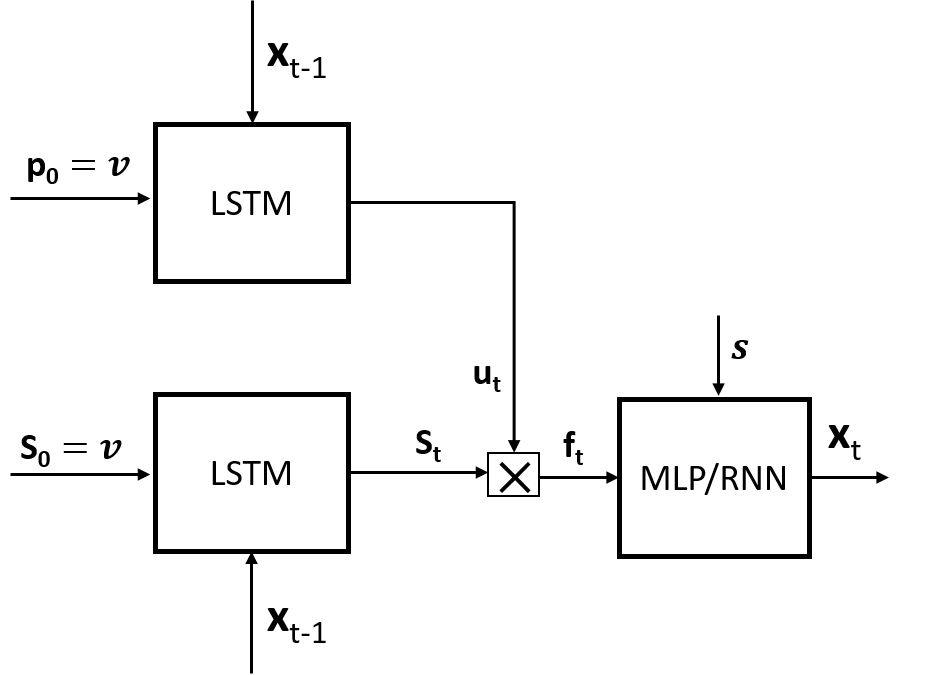}  % ,height=\textheight,keepaspectratio
\caption{Open-End CRUR} \label{fig:Open-End} 
\end{figure}
\begin{figure}[H]
\centering
\includegraphics[width=.5\textwidth]{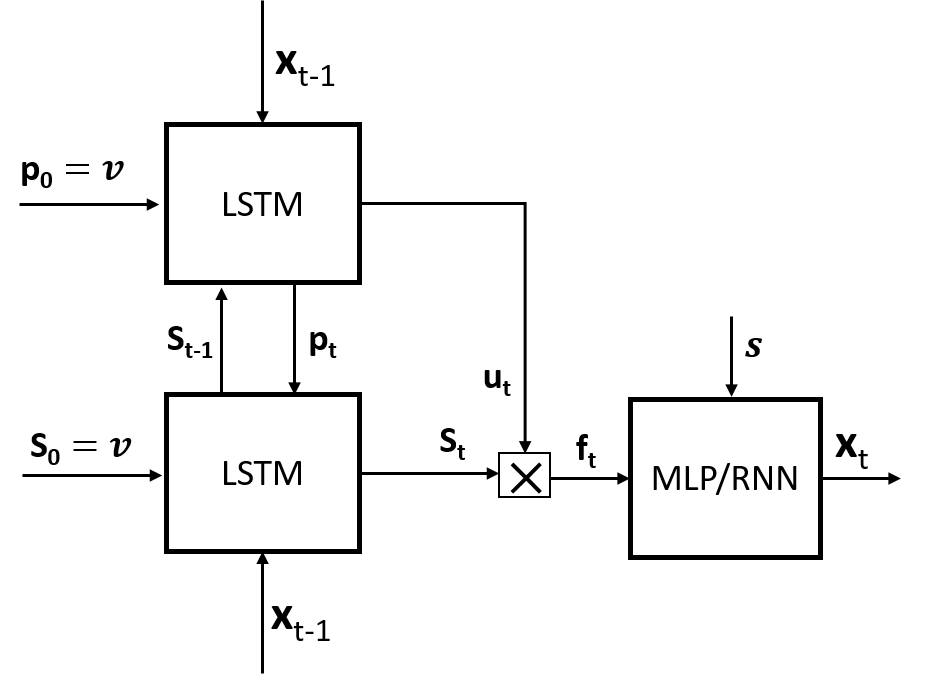}  % ,height=\textheight,keepaspectratio
\caption{Closed-End CRUR} \label{fig:Closed-End}
\end{figure}

\subsection{RNN Coupled CRUR}
Coupled RNN is based on Recurrent Neural Network units and can be regarded as most traditional unit scheme. The RNN with equations $ h_{t} = x_{t} + h_{t-1} $ is much better for applications instead of equations $ h_{t} = x_{t} + y_{t-1} $ because the former has added advantage of dependency on previous latent state space than the dependency on the previous output space. Initialization of state variables like $\textbf{S}_t$ and $\textbf{p}_t$ is important for sequential learning. While $\textbf{S}_0$ and $\textbf{p}_0$ is initialized randomly  with zero or very low range $[0.001,-0.001]$ tensors using constant seed. The initial $\textbf{S}_t$ and $\textbf{p}_t$ for the model may even get the touch of visual features to align itself with the contextual information like ($\textbf{S}_t = \textbf{W}_S \textbf{v}$ and $\textbf{p}_t = \textbf{W}_p \textbf{v}$). The model is initialization with the followings,  
\begin{equation}
 \textbf{S}_{1} = \sigma(\mathds{1}(t=1)\textbf{v}\textbf{C}_{1,S})
\end{equation}
\begin{equation}
 \textbf{p}_{1} = \sigma(\mathds{1}(t=1)\textbf{w}\textbf{C}_{1,p})
\end{equation}
where we have $\textbf{v}$, $\textbf{w} \in \mathbb{R}^{2048}$, $ \mathbb{R}^{1000}$ as the visual features and as the semantic information for the images components respectively.

\paragraph{Coupled Open-End RNN}
Coupled-oRNN or Coupled Open-End RNN is characterized by no physical interaction between the parallel units and low interactions. This kind of phenomenon does not promote cooperation and with respect to distribution analysis, the units help in divergence of the different possibilities and thus helps in exploration, but unfortunately does not provide enough help in generation, but will provide better accuracy for supervised learning problems. However, if we do an analysis which has evaluations that measure the diversity and innovation of the generator, the Open-End version will be much better. Also, when in comes to distribution, it expands the working area, but whether it helps in generalization cannot be answered without experimentation. The iteration for generation starts with these equations.
\begin{equation}
 \textbf{S}_{t} = \sigma(\mathds{1}(t>1)\textbf{x}_{1,t-1}\textbf{D}_{1,S} + \textbf{S}_{t-1}\textbf{U}_{1,S})
\end{equation}
\begin{equation}
 \textbf{S}_{t} = RNN_S(\textbf{p}_{t-1},\textbf{p}_{t-1})
\end{equation}
\begin{equation}
 \textbf{p}_{t} = \sigma(\textbf{p}_{t-1}\textbf{W}_{1,p} + \mathds{1}(t>1)\textbf{x}_{2,t-1}\textbf{D}_{1,p})
\end{equation}
\begin{equation}
 \textbf{S}_{t} = RNN_S(\textbf{p}_{t-1},\textbf{p}_{t-1})
\end{equation}
where we have the same nomenclature as the LSTM network detailed below.

\paragraph{Coupled Closed-End RNN}
Coupled-cRNN or Coupled Closed-End RNN has the inter-connectivity among the hidden states and is marked by the exchange of the hidden states where the previous states of one unit help both the units to predict the next one. This model will be marked by the stability of the generation with high accuracy, but less exploration and dynamics of the model. However, when it comes to reoccurred of what has been learned by the model, this model have tendency of regeneration of those sequence and thus helps in better distribution modeling of the working space. When it comes to risk management of the working domain, this model will have much more stability in inference and what is being taught to learn. The generation iteration framework works on these following equations. 
\begin{equation}
 \textbf{S}_{t} = \sigma(\textbf{p}_{t-1}\textbf{W}_{1,S} + \mathds{1}(t>1)\textbf{x}_{1,t-1}\textbf{D}_{1,S} + \textbf{S}_{t-1}\textbf{U}_{1,S})
\end{equation}
\begin{equation}
 \textbf{S}_{t} = RNN_S(\textbf{p}_{t-1},\textbf{p}_{t-1})
\end{equation}
\begin{equation}
 \textbf{p}_{t} = \sigma(\textbf{p}_{t-1}\textbf{W}_{1,p} + \mathds{1}(t>1)\textbf{x}_{2,t-1}\textbf{D}_{1,p} + \textbf{S}_{t-1}\textbf{U}_{1,p})
\end{equation}
\begin{equation}
 \textbf{p}_{t} = RNN_p(\textbf{p}_{t-1},\textbf{p}_{t-1})
\end{equation}
The rest of the equation for estimation of occurrence 
\begin{equation}
 \textbf{u}_{t} = \sigma(\textbf{W}_{u}\textbf{p}_{t})
\end{equation}
\begin{equation}
 \textbf{f}_{t} = \textbf{S}_{t} \textbf{u}_{t}
\end{equation}
\begin{equation}
 \textbf{x}_{t} =   (\arg\max\sigma( \textbf{W}_{x}\textbf{f}_{t} )) \textbf{W}_e
\end{equation}
where we have $\textbf{x}_{t}$ as the embedding of the predicted segment of the sentence.

\subsection{LSTM Coupled CRUR}
Coupled LSTM consisted of LSTM units and had far reaching effects for different applications due to the domain intricacies, the model can incorporate to enhance learning capabilities. 
This kind of definition and interpretation of the variables of the model can help in better structuring and framing of sentences and can help in controlling the writing style and  sentence complexity.
When it comes to performance evaluation based on the reference sentences, this model has the highest accuracy based on different statistical models. 
Initialization and refined initialization for $\textbf{p}_0$ and $\textbf{S}_0$ as visual features $\textbf{v} \in \mathbb{R}^n$ is used as functional transformation or$f(\textbf{v})$ mainly involved in  regularization and reduction of the dimension. Coupled LSTM can also defined as an open-ended and closed-ended models. 

\paragraph{Coupled Open-End LSTM}
Coupled-oLSTM or Coupled Open-End LSTM does not share any hidden state and processed the contexts independently and this is the reason why it is more sensitive to variations and when it comes to prediction, which is based on some reference sentence, it is not as fruitful as the closed architecture. In fact, bi directional is a sister of this framework. The equations for the Coupled-oLSTM are as follows, 
\begin{equation} \label{eq:o_start1}
 \textbf{i}_{1,t} = \sigma(\mathds{1}(t>1)\textbf{x}_{1,t-1}\textbf{D}_{1,i} + \textbf{S}_{t-1}\textbf{U}_{1,i})
\end{equation}
\begin{equation}
 \textbf{f}_{1,t} = \sigma(\mathds{1}(t>1)\textbf{x}_{1,t-1}\textbf{D}_{1,f} + \textbf{S}_{t-1}\textbf{U}_{1,f})
\end{equation}
\begin{equation}
 \textbf{o}_{1,t} = \sigma(\mathds{1}(t>1)\textbf{x}_{1,t-1}\textbf{D}_{1,o} + \textbf{S}_{t-1}\textbf{U}_{1,o})
\end{equation}
\begin{equation}
 \textbf{g}_{1,t} = \sigma(\mathds{1}(t>1)\textbf{x}_{1,t-1}\textbf{D}_{1,c} + \textbf{S}_{t-1}\textbf{U}_{1,c})
\end{equation}
\begin{equation} 
\textbf{c}_{1,t}  = \textbf{f}_{1,t} \odot \textbf{c}_{1,t-1} + \textbf{i}_{1,t} \odot \textbf{g}_{1,t}
\end{equation}
\begin{equation} \label{eq:o_end1}
 \textbf{S}_t = \textbf{o}_{1,t} \odot \sigma(\textbf{c}_{1,t})
\end{equation}
where we replace Equation \ref{eq:o_start1} to Equation \ref{eq:o_end1} as the following equation.
\begin{equation}
 \textbf{S}_t = LSTM_S^o(\textbf{x}_{t-1},\textbf{S}_{t-1})
\end{equation}
Similarly, the other unit, which sometimes are regarded as the converger of the context to the most effective likelihood for better prediction, is provided as the followings.
\begin{equation} \label{eq:o_start2}
 \textbf{i}_{2,t} = \sigma(\textbf{p}_{t-1}\textbf{W}_{2,i} + \mathds{1}(t>1)\textbf{x}_{2,t-1}\textbf{D}_{2,i} )
\end{equation}
\begin{equation}
 \textbf{f}_{2,t} = \sigma(\textbf{p}_{t-1}\textbf{W}_{2,f} + \mathds{1}(t>1)\textbf{x}_{2,t-1}\textbf{D}_{2,f} )
\end{equation}
\begin{equation}
 \textbf{o}_{2,t} = \sigma(\textbf{p}_{t-1}\textbf{W}_{2,o} + \mathds{1}(t>1)\textbf{x}_{2,t-1}\textbf{D}_{2,o} )
\end{equation}
\begin{equation}
 \textbf{g}_{2,t} = \sigma(\textbf{p}_{t-1}\textbf{W}_{2,c} + \mathds{1}(t>1)\textbf{x}_{2,t-1}\textbf{D}_{2,c} )
\end{equation}
\begin{equation}
\textbf{c}_{2,t}  = \textbf{f}_{2,t} \odot \textbf{c}_{2,t-1} + \textbf{i}_{2,t} \odot \textbf{g}_{2,t}
\end{equation}
\begin{equation} \label{eq:o_end2}
 \textbf{p}_t = \textbf{o}_{2,t} \odot \sigma(\textbf{c}_{2,t})
\end{equation}
where we define Equation \ref{eq:o_start2} to Equation \ref{eq:o_end2} as the following equation.
\begin{equation}
 \textbf{p}_t = LSTM_p^o(\textbf{x}_{t-1},\textbf{p}_{t-1})
\end{equation}

\paragraph{Coupled Closed-End LSTM}
Coupled-cLSTM or Coupled Closed-End LSTM shares the intermediates and hence the effect of context initialzation is also doubled and the chances of a profound interpretation chance gets raised. A large number of applications literally depend on the initialzation of the network and this initializxatioon is interpreted as as weighted selection of some portion of the contexts that is selected heuristically and is learned with the training sequences. However, most of the time, the heuristic selection can be regarded as summation of the different segments of the contexts. The equations of the Coupled-cLSTM, which iterates through the sequence of the features, starts with the following equations.
\begin{equation} \label{eq:c_start1}
 \textbf{i}_{1,t} = \sigma(\textbf{p}_{t-1}\textbf{W}_{1,i} + \mathds{1}(t>1)\textbf{x}_{1,t-1}\textbf{D}_{1,i} + \textbf{S}_{t-1}\textbf{U}_{1,i})
\end{equation}
\begin{equation}
 \textbf{f}_{1,t} = \sigma(\textbf{p}_{t-1}\textbf{W}_{1,f} + \mathds{1}(t>1)\textbf{x}_{1,t-1}\textbf{D}_{1,f} + \textbf{S}_{t-1}\textbf{U}_{1,f})
\end{equation}
\begin{equation}
 \textbf{o}_{1,t} = \sigma(\textbf{p}_{t-1}\textbf{W}_{1,o} + \mathds{1}(t>1)\textbf{x}_{1,t-1}\textbf{D}_{1,o} + \textbf{S}_{t-1}\textbf{U}_{1,o})
\end{equation}
\begin{equation}
 \textbf{g}_{1,t} = \sigma(\textbf{p}_{t-1}\textbf{W}_{1,c} + \mathds{1}(t>1)\textbf{x}_{1,t-1}\textbf{D}_{1,c} + \textbf{S}_{t-1}\textbf{U}_{1,c})
\end{equation}
\begin{equation} 
\textbf{c}_{1,t}  = \textbf{f}_{1,t} \odot \textbf{c}_{1,t-1} + \textbf{i}_{1,t} \odot \textbf{g}_{1,t}
\end{equation}
\begin{equation} \label{eq:c_end1}
 \textbf{S}_t = \textbf{o}_{1,t} \odot \sigma(\textbf{c}_{1,t})
\end{equation}
where we represent Equation \ref{eq:c_start1} to Equation \ref{eq:c_end1} as the following equation.
\begin{equation}
 \textbf{S}_t = LSTM_S^c(\textbf{x}_{t-1},\textbf{S}_{t-1},\textbf{p}_{t-1})
\end{equation}
The parallel unit, which contributes for the structuring of the sentences or the generated sequence is provided as the followings,
\begin{equation} \label{eq:c_start2}
 \textbf{i}_{2,t} = \sigma(\textbf{p}_{t-1}\textbf{W}_{2,i} + \mathds{1}(t>1)\textbf{x}_{2,t-1}\textbf{D}_{2,i} + \textbf{S}_{t-1}\textbf{U}_{2,i})
\end{equation}
\begin{equation}
 \textbf{f}_{2,t} = \sigma(\textbf{p}_{t-1}\textbf{W}_{2,f} + \mathds{1}(t>1)\textbf{x}_{2,t-1}\textbf{D}_{2,f} + \textbf{S}_{t-1}\textbf{U}_{2,f})
\end{equation}
\begin{equation}
 \textbf{o}_{2,t} = \sigma(\textbf{p}_{t-1}\textbf{W}_{2,o} + \mathds{1}(t>1)\textbf{x}_{2,t-1}\textbf{D}_{2,o} + \textbf{S}_{t-1}\textbf{U}_{2,o})
\end{equation}
\begin{equation}
 \textbf{g}_{2,t} = \sigma(\textbf{p}_{t-1}\textbf{W}_{2,c} + \mathds{1}(t>1)\textbf{x}_{2,t-1}\textbf{D}_{2,c} + \textbf{S}_{t-1}\textbf{U}_{2,c})
\end{equation}
\begin{equation}
\textbf{c}_{2,t}  = \textbf{f}_{2,t} \odot \textbf{c}_{2,t-1} + \textbf{i}_{2,t} \odot \textbf{g}_{2,t}
\end{equation}
\begin{equation} \label{eq:c_end2}
 \textbf{p}_t = \textbf{o}_{2,t} \odot \sigma(\textbf{c}_{2,t})
\end{equation}
where we symbolize Equation \ref{eq:c_start2} to Equation \ref{eq:c_end2} as the following equation.
\begin{equation}
 \textbf{p}_t = LSTM_p^c(\textbf{x}_{t-1},\textbf{p}_{t-1},\textbf{S}_{t-1})
\end{equation}
The final representation for estimation of the prediction likelihood of a word is provided as the joint tensor derived out the product of the individuals $\textbf{S}_{t}$ and $\textbf{u}_{t}$ which can be regarded as the context predictor and the structural component predictor.
\begin{equation} \label{eq:ut}
 \textbf{u}_{t} = \sigma(\textbf{W}_{u}\textbf{p}_{t})
\end{equation}
\begin{equation}
 \textbf{f}_{t} = \textbf{S}_{t} \textbf{u}_{t}
\end{equation}
\begin{equation}
 \textbf{x}_{t} =   (\arg\max \sigma( \textbf{W}_{x}\textbf{f}_{t} )) \textbf{W}_e
\end{equation}
where $\{\textbf{x}_{1},\ldots,\textbf{x}_{n}\}$ is the generated sequence.

\subsection{GRU Coupled CRUR}
GRU Coupled CRUR is composed of  Gated Recurrent Units. While LSTM is known to have the maximum effectiveness in long memory retention, GRU is known for its ability for better prediction and likelihood estimation and hence in many applications where the prediction required the final layer or the likelihood estimation layer to be sensitive, it performs better. While, most of the work is based on estimation of the sequence quality of the sentences, we have experimented GRU to determine the position of the GRU in the hierarchy of the memory unit processing and generation capability. 
Coupled architecture with GRU processes lesser number of weight estimations than Coupled LSTM and the analysis is more focused on whether we are gaining considerably with more weights or the convergence of some of the pipelines and activation units in GRU compensates for them.

\paragraph{Coupled Open-End GRU}
Coupled-oGRU or or Coupled Open-End GRU follows the open structure principle of late fusion of the likelihood without any inter communication among the memory networks. The equations for GRU based CRUR is noted by the followings, 
\begin{equation} \label{eq:o_start3}
 \textbf{z}_{1,t} = \sigma(\mathds{1}(t>1)\textbf{x}_{1,t-1}\textbf{D}_{1,z} + \textbf{S}_{t-1}\textbf{U}_{1,z})
\end{equation}
\begin{equation}
 \textbf{r}_{1,t} = \sigma(\mathds{1}(t>1)\textbf{x}_{1,t-1}\textbf{D}_{1,r} + \textbf{S}_{t-1}\textbf{U}_{1,r})
\end{equation}
\begin{multline} \label{eq:o_end3}
\textbf{S}_{t}  = \textbf{z}_{1,t} \odot \textbf{S}_{t-1} + (1-\textbf{z}_{1,t}) \odot \\
 \tanh( \mathds{1}(t>1)\textbf{x}_{1,t-1}\textbf{D}_{1,S} + (\textbf{r}_{1,t} \odot \textbf{S}_{t-1})\textbf{U}_{1,S})
\end{multline}
where we define Equation \ref{eq:o_start3} to Equation \ref{eq:o_end3} as the following equation,
\begin{equation}
 \textbf{S}_t = GRU_S^o(\textbf{x}_{t-1},\textbf{S}_{t-1},\textbf{p}_{t-1})
\end{equation}
The other GRU unit, which learns the topological dependencies of the sequence is defined as the following equations,
\begin{equation} \label{eq:o_start4}
 \textbf{z}_{2,t} = \sigma(\textbf{p}_{t-1}\textbf{W}_{2,z} + \mathds{1}(t>1)\textbf{x}_{2,t-1}\textbf{D}_{2,z} )
\end{equation}
\begin{equation}
 \textbf{r}_{2,t} = \sigma(\textbf{p}_{t-1}\textbf{W}_{2,r} + \mathds{1}(t>1)\textbf{x}_{2,t-1}\textbf{D}_{2,r} )
\end{equation}
\begin{multline} \label{eq:o_end4}
\textbf{p}_{t}  = \textbf{z}_{2,t} \odot \textbf{p}_{t-1} + (1-\textbf{z}_{2,t}) \odot  \\
  \tanh((\textbf{r}_{2,t} \odot \textbf{p}_{t-1})\textbf{W}_{2,S} + \mathds{1}(t>1)\textbf{x}_{2,t-1}\textbf{D}_{2,S} )
\end{multline}
where we replace Equation \ref{eq:o_start4} to Equation \ref{eq:o_end4} as the following equation,
\begin{equation}
 \textbf{S}_t = GRU_p^o(\textbf{x}_{t-1},\textbf{S}_{t-1},\textbf{p}_{t-1})
\end{equation}

\paragraph{Coupled Closed-End GRU}
Coupled-cGRU or Coupled Closed-End GRU iterates aound the following set of equations are likewise exchanges information though entanglement among the memory units.
\begin{equation} \label{eq:c_start3}
 \textbf{z}_{1,t} = \sigma(\textbf{p}_{t-1}\textbf{W}_{1,z} + \mathds{1}(t>1)\textbf{x}_{1,t-1}\textbf{D}_{1,z} + \textbf{S}_{t-1}\textbf{U}_{1,z})
\end{equation}
\begin{equation}
 \textbf{r}_{1,t} = \sigma(\textbf{p}_{t-1}\textbf{W}_{1,r} + \mathds{1}(t>1)\textbf{x}_{1,t-1}\textbf{D}_{1,r} + \textbf{S}_{t-1}\textbf{U}_{1,r})
\end{equation}
\begin{multline} \label{eq:c_end3}
\textbf{S}_{t}  = \textbf{z}_{1,t} \odot \textbf{S}_{t-1} + (1-\textbf{z}_{1,t}) \odot \tanh((\textbf{z}_{1,t} \odot \textbf{S}_{t-1})\textbf{W}_{1,S} \\
+ \mathds{1}(t>1)\textbf{x}_{1,t-1}\textbf{D}_{1,S} + (\textbf{r}_{1,t} \odot \textbf{S}_{t-1})\textbf{U}_{1,S})
\end{multline}
where we represent Equation \ref{eq:c_start3} to Equation \ref{eq:c_end3} as the following equation,
\begin{equation}
 \textbf{S}_t = GRU_S^c(\textbf{x}_{t-1},\textbf{S}_{t-1},\textbf{p}_{t-1})
\end{equation}
Similarly, for closed structure, the other GRU unit that governs the grammatical and part-of-speech integrity of the   sentences is denoted as the followings, 
\begin{equation} \label{eq:c_start4}
 \textbf{z}_{2,t} = \sigma(\textbf{p}_{t-1}\textbf{W}_{2,z} + \mathds{1}(t>1)\textbf{x}_{2,t-1}\textbf{D}_{2,z} + \textbf{S}_{t-1}\textbf{U}_{2,z})
\end{equation}
\begin{equation}
 \textbf{r}_{2,t} = \sigma(\textbf{p}_{t-1}\textbf{W}_{2,r} + \mathds{1}(t>1)\textbf{x}_{2,t-1}\textbf{D}_{2,r} + \textbf{S}_{t-1}\textbf{U}_{2,r})
\end{equation}
\begin{multline} \label{eq:c_end4}
\textbf{p}_{t}  = \textbf{z}_{2,t} \odot \textbf{p}_{t-1} + (1-\textbf{z}_{2,t}) \odot \tanh((\textbf{z}_{2,t} \odot \textbf{p}_{t-1})\textbf{W}_{2,S} \\
   + \mathds{1}(t>1)\textbf{x}_{2,t-1}\textbf{D}_{2,S} + (\textbf{r}_{2,t} \odot \textbf{p}_{t-1})\textbf{U}_{2,S})
\end{multline}
where we define Equation \ref{eq:c_start4} to Equation \ref{eq:c_end4} as the following equation,
\begin{equation}
 \textbf{S}_t = GRU_p^c(\textbf{x}_{t-1},\textbf{S}_{t-1},\textbf{p}_{t-1})
\end{equation}
The final representation $\textbf{f}_{t}$ is generated as a product of the tensors and is considered as a likelihood of the next word, as $\textbf{x}_{t}$, to be predicted and is directly relative to context $\textbf{S}_{t}$ and the parts of speech component $\textbf{u}_{t}$. 
\begin{equation}
 \textbf{u}_{t} = \sigma(\textbf{W}_{u}\textbf{p}_{t})
\end{equation}
\begin{equation}
 \textbf{f}_{t} = \textbf{S}_{t} \textbf{u}_{t}
\end{equation}
\begin{equation}
 \textbf{x}_{t} =   (\arg\max \sigma( \textbf{W}_{x}\textbf{f}_{t} )) \textbf{W}_e
\end{equation}
generated at time $t$, $\{\textbf{x}_{1},\ldots,\textbf{x}_{n}\}$ is the sequence.

Lastly, it must be mentioned that overall, the main principle of the architecture is dependent on the fact that Open model promotes ``Late Fusion" of the features which has been transformed non-linearly as likelihood and thus encourages pure processing of the features. These features are sensitive to variations and are itself are in pure form and this kind of late fusion of features and representation helps in better prediction of inference. They are more favorable to decision making and less influential to generative demonstrations. On the other hand, ``Early Fusion" takes place in Closed models and hence, diverse opinions get framed at a very early part of the processing of the features. In early fusion, the chances of combination of a diverse sector of interpretation also gets enhanced and trained and thus these kinds of models are more sensitive to contexts and generation of sentences are not mere repetition of similar sequences. 
%\subsection{Coupled-cbiLSTM}
%Coupled-cbiLSTM or Coupled Closed-End biLSTM 

\subsection{Generalized and Customized Representation}
\begin{figure}[H]
\centering
\includegraphics[width=.5\textwidth]{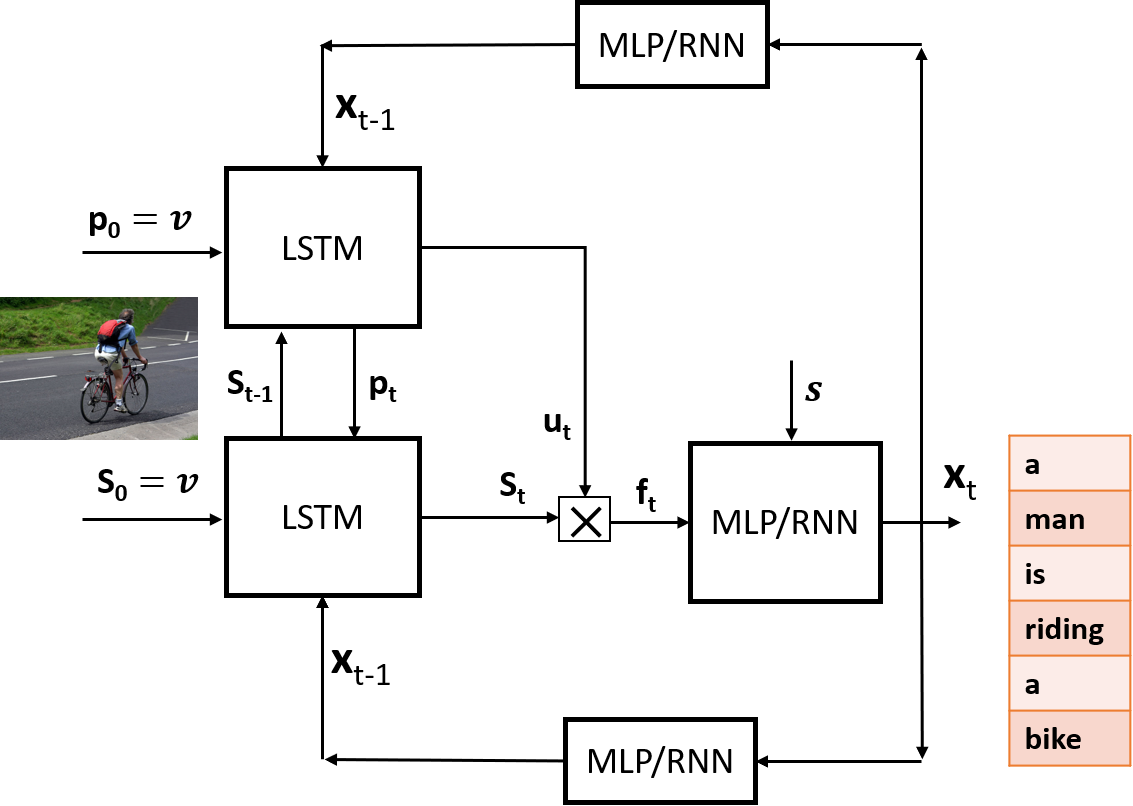}  % ,height=\textheight,keepaspectratio
\caption{Generalized (MLP) and Customized (RNN) Representation Feedback Based CRUR} \label{fig:customizedCRUR}
\end{figure}

While describing the different dual models, we realized that looping around the same kind of feedback through the model can be detrimental as each of them has different roles for accomplishment. Hence we define two other ways of feedback and provided a comparative study. 
Mathematically, the most common notion is the followings, 
\begin{equation}
 \textbf{x}_{1,t-1}, \textbf{x}_{2,t-1} = \textbf{W}_e\textbf{x}_{t-1}, \textbf{W}_e\textbf{x}_{t-1}
\end{equation}
The other two training models can be regarded as generalized feedback as the feedback learns to adapt to the changes, the model goes through. These two feedback scheme can be regarded as a customized feedback as well as it provides better smoothness for the optimization for the generation of sequential dependencies. 
Mathematically, the other two schemes, with MLP and memory respectively, can be defined as follows, 
\begin{equation}
 \textbf{x}_{1,t-1}, \textbf{x}_{2,t-1} = \textbf{W}_1\textbf{W}_e\textbf{x}_{t-1}, \textbf{W}_2\textbf{W}_e\textbf{x}_{t-1}
\end{equation}
\begin{equation}
 \textbf{x}_{1,t-1}, \textbf{x}_{2,t-1} = LSTM_1(\textbf{W}_e\textbf{x}_{t-1}), LSTM_2(\textbf{W}_e\textbf{x}_{t-1})
\end{equation}
As future work, other feedback schemes like the POS structure embedding of the sentence can be used as feedback for the system. Figure \ref{fig:customizedCRUR} provided a diagramtic overview of the two different feedback training schemes.

\section{Methodology Specifics} \label{section:methodology} %  to Application
Language attributes are different in characteristics and their prediction requires distinct modeling and interpretation while maintaining topological relationships among the different components. This section will mainly describe the details of the data, the experiments performed on them, numerical and qualitative results and interpretation of inference of the different statistical evaluations. 

\subsection{Application Description}
Image captioning is an effective way of transforming visual and media images to meaningful sentences that describe certain actions and activities, detected in the media. Though there are thousands of possibilities of such captions, it is an effort to make the machine acquainted with what the possibilities it can see in the image and can reproduce in the captions. This has immense applications ranging from detection of object activities, answering questions about images and videos, story narration and commenting about the events in videos. In this work, we have focused on the prospect of language attribute control for the caption generator application.

\subsection{Dataset Preparation}
MSCOCO dataset has been used for our experiments and has undergone much data engineering due to the involvement of immense interest and a large community from industry.   
MSCOCO consists of $123287$ training images and 566747 training sentence, where each image is associated with at least five sentences from a vocabulary of 8791 words. There are 5000 images (with 25010 sentences) for validation and 5000 images (with 25010 sentences) for testing.
Each visual images corresponds to at least 5 different sentences, creating a pool of around (566+25+25)K sentences with vocabulary of $19K$. A large part of the words were under-represented and \cite{Gan2016} used a total of $8.7K$ words for the training.  
ResNet features description are used for visual images through transferring the learned knowledge from already trained Residual Network.
Two set of features are being widely used for fusion, image features and the probability of the highly occurring objects in images, where ResNet features consists of $2048$ dimension feature vector while the other is Tag features with feature vector of $999$ dimension.

\subsection{Different Tensor Regularization}
The dimension of Tensor must be considerable. Effective learning happens when the hidden layer dimension is optimum, which can be difficult to define. Hence proper regularization is necessary for many variables. We have used dropout value of $0.5$ for word embedding and the generated TPR tensor denoted as $\textbf{f}_t$. For effective decoding of the captions, an effective learning of the word embedding matrix $W_e$ is necessary along with a layer of Tensor Regularization through dropout. Magically, $0.5$ dropout rate happened to be better accuracy than $1$, $0.7$ and $0.4$ and it is difficult to figure out the reason. 

Different features merge into the memory network to compose a perfect and favorable composition called representation that can be identified and decoupled into a sentence and a significant amount of dropout is required for each of them. A common dropout at the entry point of the network had worked but it is not enough as the combinational effect of each of them is selected and generalized. Different dropout combinations will not only help in proper selection and estimation of parameters but also help in generation of unique combination for different sequences. It must be noted that optimum amount of dropout rate is kept at $0.5$ because of the fact that $0.5$ helps in protecting at least more than $50\%$ of the feature participation for the tensors and thus scarcity of important contents is prevented while training the model. Also, $0.5$ helps in preventing over dominance of features and thus help in creation of diversity in the representation. It also makes the model sensitive to the variations of the features. 

\subsection{Normalization of Images Features}
In our experiments, we observed that the individual  normalization of the ResNet visual features with mean of the features vector can enhance the BLEU\_4 accuracy of the model with an improvement of $(100\times 1/27 = 3.7 \%)$, which will ultimately enhance the overall context visualization situation for the model. This will bound the features to range of $\mathbb{R}^{2048} \in \mathcal{N}(\mu = 0,\,\sigma^{2})$ in comparison to the previous $\mathbb{R}^{2048} \in \mathcal{N}(\mu \neq 0,\,\sigma^{2})$. This is a significant event for the diverse representation of the image features, mainly when we are defining that suppression of the variations is not a good phenomenon in generative models as it reduces the sensitivity of the model to variations. However, it is fine as far as effectiveness is going high and the feature variations are suppressed at a very low level. Nevertheless, it must be mentioned that the most effectiveness of the handling features by a model is through normalization and had been well established fact in statistics. We have performed individual normalization and is noted mathematically as the following,
\begin{equation}
 \overline{\textbf{v}} = \textbf{v} - \frac{1}{n}\sum\limits_i^n v_i  
\end{equation}
where $v_i \in \{v_1,v_2,\ldots,v_n\} \in \textbf{v}$. 

\subsection{Normalization of Word Vector}
Global vector for all machines, just like vocabulary can create generalization and cross-system interpretation possibility. However, if the global vocabulary representations are used like those provided by Word2Vec, GloVe etc, the effectiveness of the model gets enhanced if the whole set of representation is normalized. Normalization of the Word Vector helped in 1.5\% improvement in BLEU\_4 accuracy which is an improvement of $(100\times 1.5/27 = 5.56 \%)$ improvement. Each word embedding vector $(\textbf{w}_e)_i$ is normalized as the following,
\begin{equation}
 (\overline{\textbf{w}}_e)_i = (\textbf{w}_e)_i - \frac{1}{(V*d)}\sum\limits_i^V \sum\limits_j^d (W_e)_{ij}  
\end{equation}
where, $W_e \in \mathbb{R}^{V\times d}$ is the word embedding matrix,  $V$ is the vocabulary length and $d$ is the dimension of the continuous representation embedding for the words based on context. The final $\overline{\textbf{w}}_e \in \mathbb{R}^d \in \mathcal{N}(\mu=0,\,\sigma^{2})$ in comparison to ${\textbf{w}}_e \in \mathbb{R}^d \in \mathcal{N}(\mu\neq 0,\,\sigma^{2})$. Global vector representation will facilitate communication among different models and create feasible opportunities for machines in term of interpretation, storage and retrieval. However, locally trained embedding vector may be better for some cases of prediction and improve by 1\% accuracy for BLEU\_4 metrics.

\subsection{Beam Search}
Beam Search provides the necessary search procedure considering variation of the prediction of the model and scope of error in sequential learning and prediction estimation and considered metrics. 
Beam Search works on the principle of expanding and trimming of the search space based on the evaluation criteria. Here the log of centered and scaled probability distribution \cite{Gan2016} of the softmax layer is used for evaluation. 
Beam Search helped in 8.93\% improvement in BLEU\_4 accuracy which is an improvement of $(100\times 2.5/28.0 = 8.93 \%)$ improvement.
Mathematically, beam search can be denoted as the following set of equations, 
\begin{equation} 
\begin{split}
p & (\theta_1,\ldots,\theta_n|\textbf{v})   \\
\propto  & \max \sum  p(\theta_i|\theta_{i-1}\ldots\textbf{v}) \\
\propto  & \max \sum  p(\theta_i|p(\theta_{i-1}|\theta_{i-2}\ldots\textbf{v})) \\
\propto  & \max \sum  p(\theta_i|p(\theta_{i-1}|p(\theta_{i-2}|\ldots p_0(\theta_1|\textbf{v}))) \\
\end{split}
\end{equation}
where $\textbf{v}$ is the extracted visual feature and $\{\theta_1,\ldots,\theta_n\}$ is the generated caption. Involving $\theta$ vocabulary set and $Sp$ as the special character set to indicate the start and end of sentence, maximum likelihood estimated as $\max p (\theta_1,\ldots,\theta_n|\textbf{v})$ $\forall$ $\theta_i \not\in Sp $ for the high value of $p(\theta_i \in Sp|\theta_{i-1}\ldots\textbf{v})$ will prune useful nodes in beam search tree to create the most probable sentence. 
Researchers has claimed ensemble to be efficient, only when the level of noise in the model is very high and variance is affecting the output. Our model didn't gain much out of  the ensemble of different model outputs. The main reason is that the probability of occurrence for $\{\theta_1,\ldots,\theta_n\}$ is varying and shifting, centering created different numerical range for different predictors with different weights. The training is mainly aimed at occurrence of truth at maximum.

\subsection{Training Procedure}
Supervised training for recurrent neural network is done mainly through feedback of the previous state(s) of the model like $\textbf{x}_{t-1} \in \mathbb{R}_{W_E}$ where $W_E$ is the dimension of the word embedding. $\textbf{x}_{t-1}$ is the real situation state output, but for better influence and better establishment of the sequential topology and low training error rate during learning phases, $\tilde{\textbf{x}}_{t-1}$ comes from the data and replaces $\textbf{x}_{t-1}$. 
In sequential learning, $\tilde{\textbf{x}}_{t-1}$ supervise the learning on the assumption that the learning is going well which ensures the long term learning structure is stable and concrete. During initial phases, when error rate is high, $\textbf{x}_{t-1}$ may be $\tilde{\textbf{x}}_{t-1}$ or may be $\textbf{x}_{error}$, but the feedback is return as $\tilde{\textbf{x}}_{t-1}$ where $\tilde{\textbf{x}}_{t-1} = \textbf{x}_{t-1}$ and $\tilde{\textbf{x}}_{t-1} \neq \textbf{x}_{t-1}$ respectively. Knowledge, in raw form from data, generates the subspace for learning while the expansion in variance of subspace is limited and thus creating bias.  To increase the variance, error is added or regularized to increase the influence of the important variables. 
This phenomenon of supervised training creates a new environment, which is different from testing, sometimes termed as drifting and is an issue for sequential learning. Drifting shifts the learning experience biased away from the visual context.
Constant injection of $\tilde{\textbf{x}}_{t-1}$ inhibits learning as the gradient diminishes and decrease in error stagnates. So we use $\textbf{x}_{t-1}$ instead of $\tilde{\textbf{x}}_{t-1}$ in some cases and this can create jerks or changes in the weights which will again try to reach a stable state. 
This problem can be acknowledge through the concept of jitter introduction. Jitter is some kind of noise which is expected during the normal progress of the operations. Introduction of jitter helps create a robust system while at the same time will prevent over-fitting. This concept is similar to the simulated annealing where diminished gradient is revived through phase transform which is one way to escape local optimum towards global one. The percentage of jitter must be very small compared to the normal training.

\subsection{TPR Attention}
TPR Attention is provided when we use the LSTM (or RNN) as decoder instead of the MLP, as shown in Figure \ref{fig:Closed-End} and the performance evaluation is shown in Table \ref{table:table1} as LSTM CRUR Attn. With just image features, this is perhaps the best possible performance (BLEU\_4 = .307) achieved so far, while these other similar performances used other kinds of features like semantic tag features etc.

\subsection{Reinforcement Learning Through SCST}
\begin{figure*}%[!h]
\centering
\includegraphics[width=\textwidth]{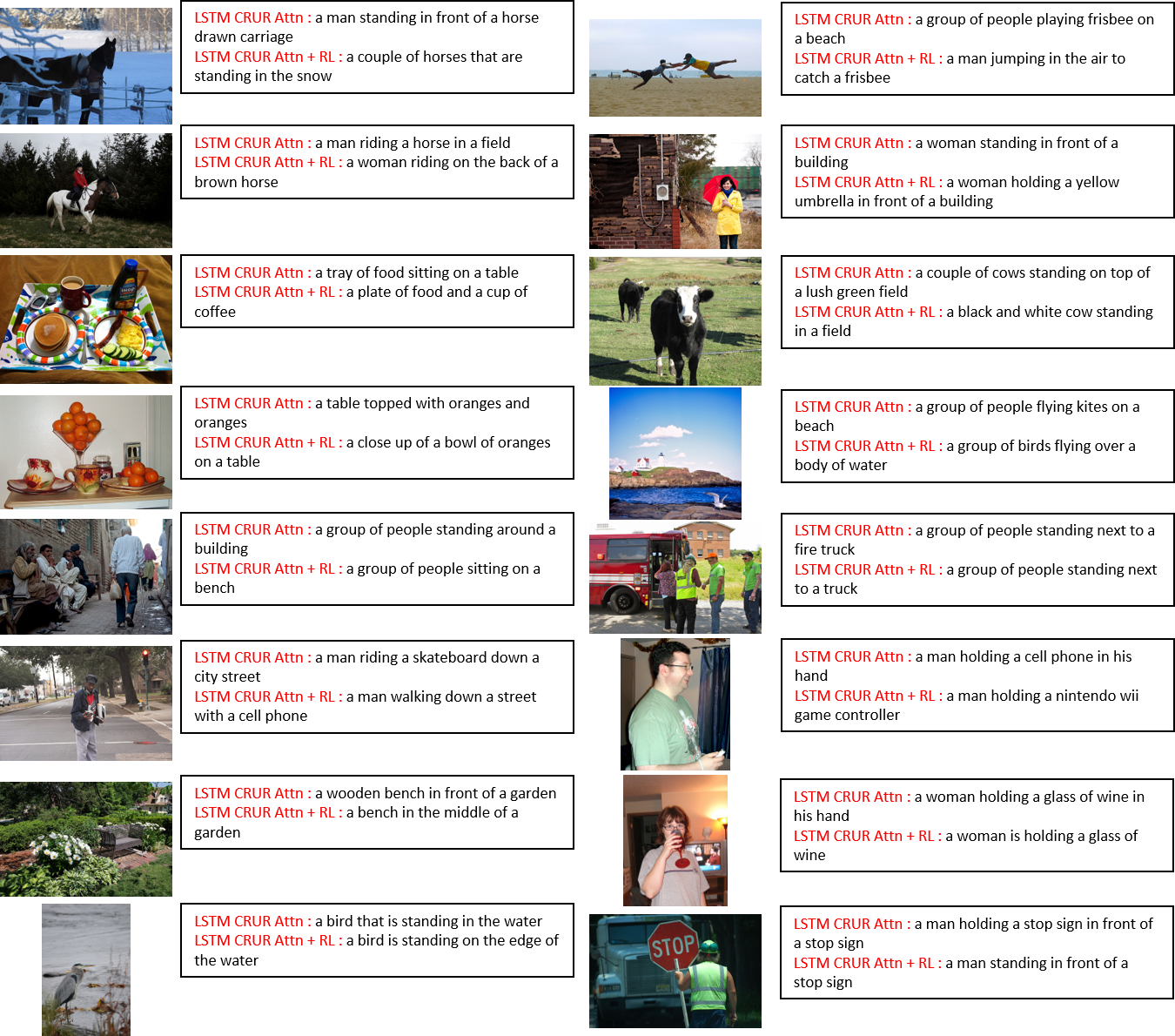}  % ,height=\textheight,keepaspectratio
\caption{Qualitative Comparison Between (LSTM CRUR Attn) and (LSTM CRUR Attn + RL)} \label{fig:ComparisonRL}
\end{figure*}

Self-critical Sequence Training (SCST) \cite{rennie2017self} works very well for image captioning applications, where the sequential dependencies help in providing an opportunity for enhancement in learning of the parameters. We achieved a performance of (BLEU\_4 = 0.327), whhen we used SCST with CIDEr-D as evaluation function for gradient feedback. We have denoted the result as LSTM CRUR Attn + RL in Table \ref{table:table1}. Figure \ref{fig:ComparisonRL} provided some improved instances of generated captions with attention model of CRUR and enhancement with RL. 
SCST based reinforcement learning can be represented as, 
\begin{equation} \label{eq:CR1}
 \frac{\delta L(\textbf{w})}{\delta \textbf{w}} = -\frac{1}{2b}\gamma \sum\limits_i \Phi(\textbf{y},\textbf{y}')
\end{equation}
\begin{equation} \label{eq:IR}
 \frac{\delta L(\textbf{w})}{\delta \textbf{w}} = -\frac{1}{2b}\gamma \sum\limits_i \Phi(\{y_1,\ldots,y_c\}, \{y'_1,\ldots,y'_c\})
\end{equation}
where $\Phi(.)$ is the evaluation function or the reward function that evaluates certain aspects of the generated captions $\{y_1,\ldots,y_c\} \in \textbf{y}$ and the baseline captions $\{y'_1,\ldots,y'_c\} \in \textbf{y}'$ and $b$ is the mini-batch size considered.

\section{Results \& Analysis} \label{section:results}
This section is focused with the results on the architecture and its performance in comparison with other LSTM and bi-LSTM architectures. CRUR is a generative network and hence we concentrated our focus on generational criteria than mere prediction evaluations and its ability to guess the correct category. In fact, CRUR requires two types of compositional ingredients to be able to logically infer that both are participating in driving the generation and there are open scope to drive the sequential prediction with innovative sentences.  

\subsection{Assessment Procedures}
Assessment criteria is diversifies through a series of statistical criteria for natural languages as a single evaluation will never be able to judge the compositional ability of the model network in establishing the topological dependency of words and parts-of-speeches into a sentence. Bleu\_n calculates the statistical average of number of combined $n$ series of words that appear in the generated sentence compared to the original sentence. Other evaluation procedures include METEOR, ROUGE\_L, CIDEr-D and SPICE and mainly measure the overall sentence fluency. 

\subsection{Quantitative Analysis} \label{subsec:Quantitative}
This part will mainly discuss the quantitative analysis for different architectures and based on different initialization with visual features. 
Table \ref{table:table1} and Table \ref{table:table2} provided a comparative study table for our architecture based on different initialization.
From the results, we can clearly say that our model performed much better than the existing architectures and have promising prospects.
In these experiments, we have used the training, validation and test set of \cite{Gan2016}, which mainly follows Karpathy's split and uses a 2048 dimension layer of ResNet101 as visual features. 

\begin{table*}
\centering
\caption{Performance Evaluation for Closed Model}
\begin{tabular}{|c|c|c|c|c|c|c|c|c|}
\hline
 Algorithm & CIDEr-D & Bleu\_4 & Bleu\_3 & Bleu\_2 & Bleu\_1 & ROUGE\_L & METEOR  &  SPICE \\ 
\hline \hline
 %RNN CRUR & & & & & & & &   \\ \hline
    Human \cite{wu2017image} & 0.85 & 0.22 & 0.32 & 0.47 & 0.66 & 0.48 & 0.2 & -- \\ 
    Neural Talk \cite{Karpathy2015Deep} & 0.66 & 0.23 & 0.32 & 0.45 & 0.63 & -- & 0.20 & --  \\ 
    Mind’sEye \cite{Chen2015Mind} & -- & 0.19 & -- & -- & -- & -- & 0.20 & --  \\
    Google \cite{vinyals2015show}  & 0.94 & 0.31 & 0.41 & 0.54 & 0.71 & 0.53 & 0.25 & --  \\
    LRCN \cite{Donahue2015Long-term} & 0.87 & 0.28 & 0.38 & 0.53 & 0.70 & 0.52 & 0.24 & --  \\   
    Montreal \cite{Xu2015Show} & 0.87 & 0.28 & 0.38 & 0.53 & 0.71 & 0.52 &  0.24 & -- \\ 
    m-RNN \cite{Mao2014deep}  & 0.79 & 0.27 & 0.37 & 0.51 & 0.68 & 0.50 & 0.23 & --  \\
    \cite{Jia2015}  & 0.81 & 0.26 & 0.36 & 0.49 & 0.67 & -- & 0.23 & --  \\
    MSR \cite{Fang2015captions} & 0.91 & 0.29 & 0.39 & 0.53 & 0.70 & 0.52 & 0.25 & --  \\
    %Xu.et.al & -- & 0.25 & 0.36 & 0.50 & 0.72 & -- & 0.23 & --  \\
    \cite{Jin2015Aligning} & 0.84 & 0.28 & 0.38 & 0.52 & 0.70 & -- & 0.24 & --  \\
    bi-LSTM \cite{wang2018image} & -- & 0.244 & 0.352 & 0.492 & 0.672 & -- & -- & --  \\ 
    MSR Captivator \cite{Devlin2015Language} & 0.93 & 0.31 & 0.41 & 0.54 & 0.72 & 0.53 & 0.25 & -- \\
    Nearest Neighbor \cite{devlin2015exploring} & 0.89 & 0.28 & 0.38 & 0.52 & 0.70 & 0.51 & 0.24 & -- \\
    MLBL \cite{kiros2014multimodal} & 0.74 & 0.26 & 0.36 & 0.50 & 0.67 & 0.50 & 0.22 & -- \\
    ATT \cite{You2016Image} & 0.94 & 0.32 & 0.42 & 0.57 & 0.73 & 0.54 & 0.25 & -- \\
    \cite{wu2017image} & 0.92 & 0.31 & 0.41 & 0.56 & 0.73 & 0.53 & 0.25 & -- \\
    LSTM-R \cite{Gan2016} & 0.889 & 0.292 & 0.390 & 0.525 & 0.698 & -- & 0.238 & --  \\ \hline \hline 
 %bi-LSTM \cite{Wang:2018} & -- & 0.244 & 0.352 & 0.492 & 0.672 & -- & -- & --  \\ \hline
 %Include other results in \cite{Gan2016} & & & & & & & & \\ \hline
 % {'CIDEr': 0.8445485992876441, 'Bleu_4': 0.2899939535675805, 'Bleu_3': 0.3924295187280933, 'Bleu_2': 0.5268277878254184, 'Bleu_1': 0.6899945086865259, 'ROUGE_L': 0.5119436097464874, 'METEOR': 0.22786463428002535, 'SPICE': 0.15640687077276375}
 LSTM CRUR + $\textbf{p}_0$ Init & 0.845 & 0.290 & 0.392 & 0.527 & 0.690 & 0.512 & 0.228 & 0.156  \\ %\hline
 %{'CIDEr': 0.8404213128624107, 'Bleu_4': 0.287371836522648, 'Bleu_3': 0.3873708786745301, 'Bleu_2': 0.5221124539338396, 'Bleu_1': 0.6906280956165599, 'ROUGE_L': 0.5113489714163814, 'METEOR': 0.22695199974532204, 'SPICE': 0.15432666960359276} 
 GRU CRUR + $\textbf{p}_0$ Init & 0.840 & 0.287 & 0.387 & 0.522 & 0.691 & 0.511 & 0.227 & 0.154  \\ %\hline
 %bi-LSTM CRUR + S Init & & & & & & & &   \\ \hline
 %LSTM CRUR + S + Tag & & & & & & & &   \\ \hline 
 %GRU CRUR + Tag & & & & & & & &   \\ \hline
 %bi-LSTM CRUR + Tag & & & & & & & &   \\ \hline
 % {'CIDEr': 0.8562996934480481, 'Bleu_4': 0.2936455569653512, 'Bleu_3': 0.39093990505439324, 'Bleu_2': 0.5233455156390883, 'Bleu_1': 0.6896689835001532, 'ROUGE_L': 0.5099282409520708, 'METEOR': 0.2289880340598457, 'SPICE': 0.1551563168332822}
 LSTM CRUR & 0.860 & 0.294 & 0.391 & 0.523 & 0.690 & 0.510 & 0.229 & 0.155  \\ %\hline 
 
 %{'CIDEr': 0.8077186454279729, 'Bleu_4': 0.2733483392932721, 'Bleu_3': 0.3696182172802732, 'Bleu_2': 0.49849874955466267, 'Bleu_1': 0.659064790108268, 'ROUGE_L': 0.49909674435932866, 'METEOR': 0.22013754754341952, 'SPICE': 0.1533728025027618}
 GRU CRUR & 0.808 & 0.273 & 0.370 & 0.499 & 0.660 & 0.499 & 0.220 & 0.153  \\ %\hline
 
 % {'CIDEr': 0.9267942410351355, 'Bleu_4': 0.30653601766569955, 'Bleu_3': 0.4070659046462367, 'Bleu_2': 0.5424537770223095, 'Bleu_1': 0.7105611481310263, 'ROUGE_L': 0.5276700846991472, 'METEOR': 0.2450966930289804, 'SPICE': 0.17537880698724212}
 LSTM CRUR Attn$\dagger$  & 0.927 & 0.307 & 0.407 & 0.542 & 0.711 & 0.528 & 0.245 & 0.175 \\ %\hline
 
 % {'CIDEr': '0.9875989254201578', 'Bleu_4': 0.32664427827000056, 'Bleu_3': 0.42966670107751986, 'Bleu_2': 0.5668908185736375, 'Bleu_1': 0.7317494848495708, 'ROUGE_L': '0.5383427675304181', 'METEOR': 0.2518315215156282, 'Spice': '0.18181969131278375'} ###### gamma was 0.5 #######
 LSTM CRUR Attn$\dagger$ + RL  & 0.988 & 0.327 & 0.430 & 0.567 & 0.732 & 0.538 & 0.252 & 0.182 \\ \hline 
 \multicolumn{3}{l}{\textsuperscript{$\dagger$}\footnotesize{Attn $\rightarrow$  LSTM in place of ``MLP/RNN'' in Figure \ref{fig:basicTPRa} }}
\end{tabular}
\label{table:table1}
\end{table*}
\begin{table*}
\centering
\caption{Performance Evaluation for Open Model}
\begin{tabular}{|c|c|c|c|c|c|c|c|c|}
\hline
 Algorithm & CIDEr-D & Bleu\_4 & Bleu\_3 & Bleu\_2 & Bleu\_1 & ROUGE\_L & METEOR  &  SPICE \\ 
\hline \hline
 %RNN CRUR & & & & & & & &   \\ \hline
 LSTM-R \cite{Gan2016} & 0.889 & 0.292 & 0.390 & 0.525 & 0.698 & -- & 0.238 & --  \\ %\hline 
 bi-LSTM \cite{wang2018image} & -- & 0.244 & 0.352 & 0.492 & 0.672 & -- & -- & --  \\ \hline \hline 
 % {'CIDEr': 0.8002659989037291, 'Bleu_4': 0.25386837977159693, 'Bleu_3': 0.3570483639633674, 'Bleu_2': 0.49366359321927455, 'Bleu_1': 0.6630538795939165, 'ROUGE_L': 0.46209687006063177, 'METEOR': 0.21729170055286096, 'SPICE': 0.15739305445565924}
 LSTM CRUR & 0.800 & 0.254 & 0.357 & 0.494 & 0.663 & 0.462 & 0.217 & 0.157  \\ %\hline 
 % {'CIDEr': 0.6472613158449946, 'Bleu_4': 0.21964567903997118, 'Bleu_3': 0.31084505190773787, 'Bleu_2': 0.4354577415110018, 'Bleu_1': 0.5956655722698712, 'ROUGE_L': 0.42217499429387656, 'METEOR': 0.19397925211579348, 'SPICE': 0.13650983271365058}
 GRU CRUR & 0.647 & 0.220 & 0.311 & 0.435 & .596 & 0.422 & 0.194 & 0.137  \\ \hline
 %bi-LSTM CRUR & & & & & & & &   \\ \hline
 %LSTM CRUR + S & & & & & & & &   \\ \hline 
 %GRU CRUR + S Init & & & & & & & &   \\ \hline
 %bi-LSTM CRUR + S Init & & & & & & & &   \\ \hline
 %LSTM CRUR + S + Tag & & & & & & & &   \\ \hline 
 %GRU CRUR + Tag & & & & & & & &   \\ \hline
 %bi-LSTM CRUR + Tag & & & & & & & &   \\ \hline
 %\multicolumn{3}{l}{\textsuperscript{**}\footnotesize{xxxx $\rightarrow$  xxxx}}
\end{tabular}
\label{table:table2}
\end{table*}

\subsection{Qualitative Analysis} \label{subsec:Qualitative}
\begin{figure*}[!h]
\centering
\includegraphics[width=\textwidth]{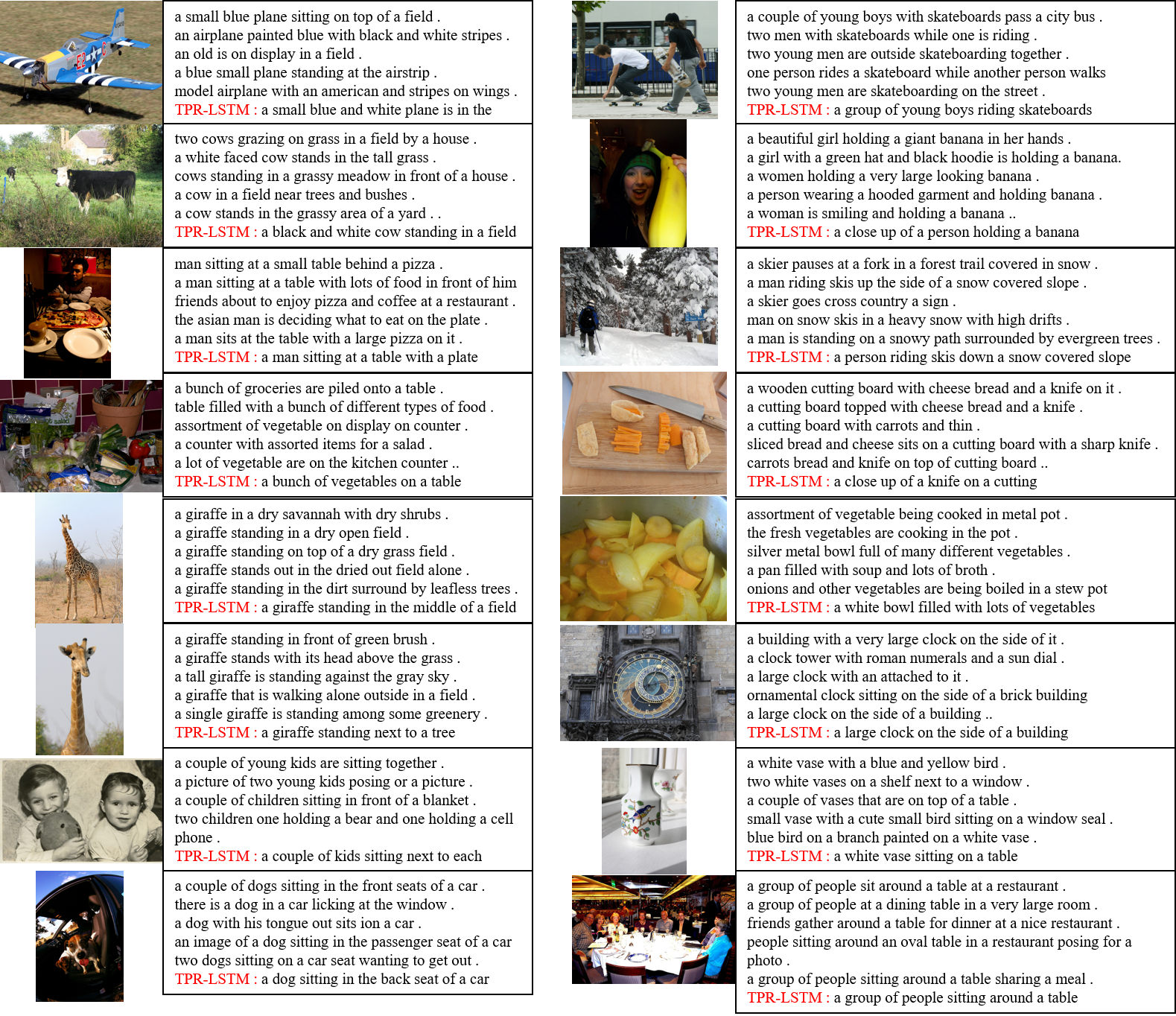}  % ,height=\textheight,keepaspectratio
%\caption{Topological Effects of Semantic Tensor $\textbf{u}_t$} \label{fig:LanguageSemanticsTopology}
\caption{Qualitative Analysis (TPR-LSTM denotes LSTM-CRUR Closed Model). Part 1.} \label{fig:QualitativeAnalysis1}
\end{figure*}

\begin{figure*}%[!h]
\centering
\includegraphics[width=\textwidth]{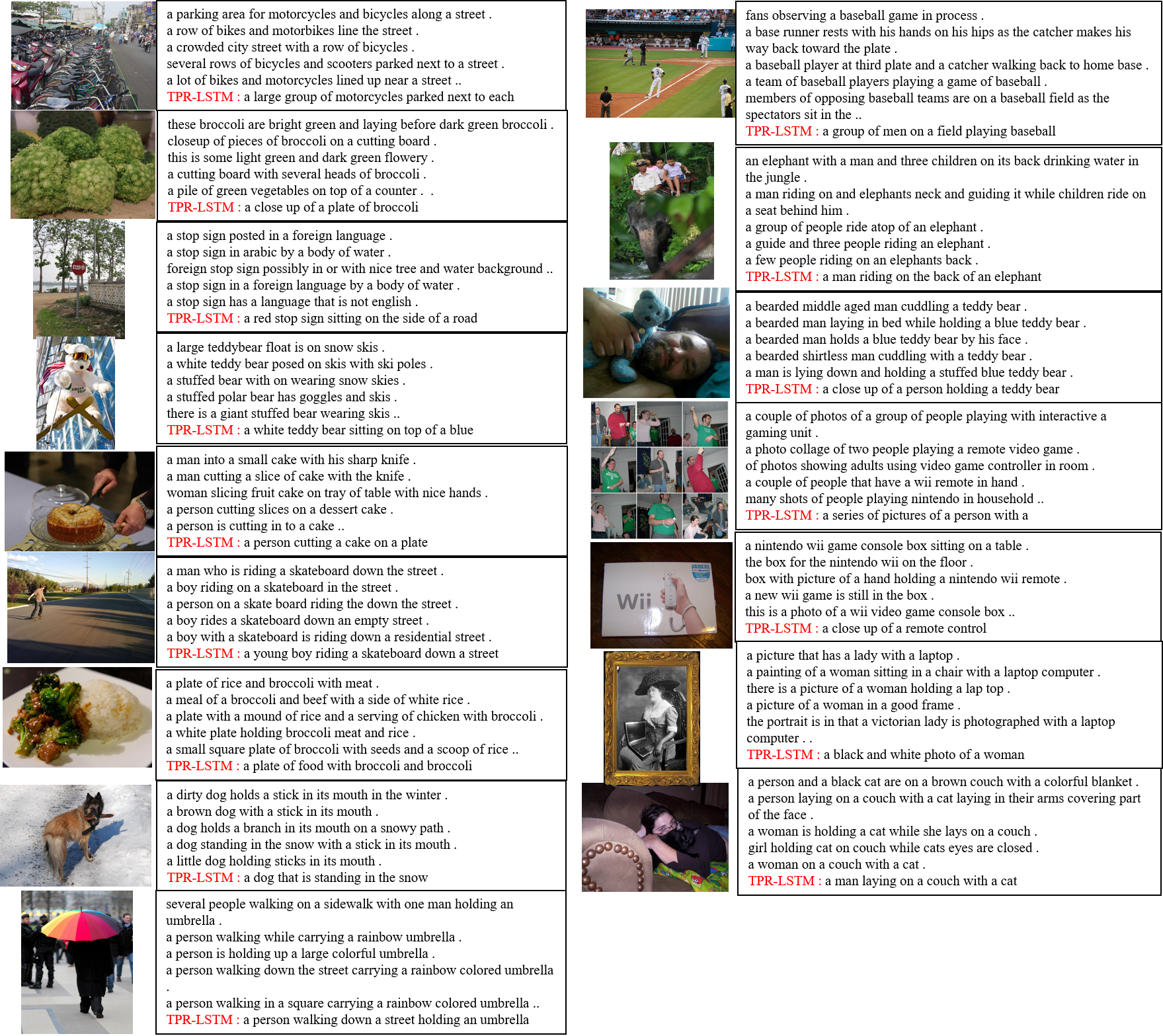}  % ,height=\textheight,keepaspectratio
%\caption{Topological Effects of Semantic Tensor $\textbf{u}_t$} \label{fig:LanguageSemanticsTopology}
\caption{Qualitative Analysis (TPR-LSTM denotes LSTM-CRUR Closed Model). Part 2. } \label{fig:QualitativeAnalysis2}
\end{figure*}

Quantitative never provides the best and aesthetic picture of languages and hence we borrowed qualitative analysis for evaluations. Here are some of the comparisons of the instances of different generated sentences from contexts in Figure \ref{fig:QualitativeAnalysis1} and in Figure \ref{fig:QualitativeAnalysis2}. Our new approach has produced much better and closely related captions for the images compared to the baseline captions. These generated captions are evidences that the architectures produce captions with novel compositions. This is also evident from the fact that the similarity is accounted with 35\% with the original reference set sentences.

%Based on the diagram in Figure \ref{fig:LanguageControling}, we provide the equations for the training of network and the control. The equations are as follows,  

\section{Discussion} \label{section:discussion}
In this work, we discussed the theoretical aspect of coupling different models, each representing different aspects likelihood and the joint fusion of the network will help in establishing the most effective likelihood of grammatically correct sequence of words as a sentence. This model is the generalization of the bi directional LSTM and provides much better insight of the architecture and their utility. Previous approach to deep learning architecture was limited to utilization and managing through evaluation of the end likelihood, an effort leading to no-understanding of the principles of why these architectures were made. However, in this work, we have discussed the different theoretical aspects that lead to an effective learning algorithm, mainly when handling fusion of different parallel architectures and sequence of topologically dependent data. While CRUR model is marked by its ability to learn different interpret-able aspects of the data, it gets rid of the pre-assumptions considered by bi-directional LSTMs, which are either misunderstood or not properly explained and documented and largely neglected.
Some of the key take away points that can be said after these analysis:
\begin{enumerate}
	\item We analyzed dual unit architecture and generalized the notion of product of tensors for exploratory generation.
	\item The tensor products help the most for learning of the language attributes of the sentences simultaneously through representation that is different from the likelihood of prediction for a word.
	\item We have done elaborated analysis of the language attributes and also came up with the controlling factor that help select sentence construction techniques, which was previously never tried before. 
	\item It is believed that detection will only be useful if we can use it for control. In that sense, we offered an approach to control different sentence constructions through likelihood of the next possible pasts-of-speech (and can be extended to simple, complex and compound). 
\end{enumerate}

\ifCLASSOPTIONcaptionsoff
  \newpage
\fi

\section*{Acknowledgments}
The author has used University of Florida HiperGator, equipped with NVIDIA Tesla K80 GPU,  extensively for the experiments. The author acknowledges University of Florida Research Computing for providing computational resources and support that have contributed to the research results reported in this publication. URL: http://researchcomputing.ufl.edu


%\end{IEEEbiography}


%\newpage



%\vfill


\end{document}